\documentclass[manuscript]{acmart}
\usepackage{natbib}
\usepackage{hyperref}
\usepackage{lscape}
\usepackage{graphicx}
\usepackage{adjustbox}
\usepackage{xcolor}
\usepackage{tikz}

\title{Multi-Step Reasoning with Large Language Models, a Survey}

\author{Aske Plaat} \author{Annie Wong} \author{Suzan Verberne} \author{Joost Broekens} \author{Niki van Stein} \author{Tho\-mas B{\"a}ck} \affiliation{\department{LIACS} \institution{Leiden University} \country{Netherlands}}

\begin{document}


\begin{abstract} 
Large language models (LLMs) with billions of parameters exhibit in-context learning abilities, enabling few-shot learning on tasks that the model was not specifically trained for. Traditional  models achieve breakthrough performance on language tasks, but do not perform well on basic reasoning benchmarks. 
However, a new in-context learning approach, Chain-of-thought, has demonstrated strong multi-step 
reasoning abilities on these benchmarks. 

The research on LLM reasoning abilities started with the question whether LLMs can solve grade school math word problems, and has expanded to other tasks in the past few years. This article reviews the field of multi-step 
reasoning with LLMs. We propose a taxonomy that identifies different ways to generate, evaluate, and control multi-step reasoning. We provide an in-depth coverage of core approaches and open problems, and we propose a research agenda for the near future. 

We find that multi-step reasoning approaches have progressed beyond math word problems, and can now successfully solve challenges in logic, combinatorial games, and robotics, sometimes by first generating code that is then executed by external tools. Many studies in multi-step methods  use reinforcement learning for finetuning, external optimization loops,  in-context reinforcement learning, and  self-reflection.


\end{abstract}

\ccsdesc[500]{Computing methodologies~Natural language processing} 

\maketitle

\section{Introduction}
Transformer-based Large Language Models (LLMs) that are trained on large datasets have achieved breakthrough performance at text generation tasks that directly build on next token prediction \citep{vaswani2017attention,radford2019language,wei2022emergent}; they are  good at natural language understanding (GLUE, SQUAD, Xsum) \citep{wang2018glue,wang2019superglue,rajpurkar2016squad,narayan2018don}, translation \citep{kocmi2022findings,papineni2002bleu,sennrich2015improving}, question answering \citep{tan2023can}, and other language generation tasks. 
The success of models such as ChatGPT \citep{ouyang2022training} is impressive.

Transformer-based generative language models whose size is beyond hundreds of billions parameters are not only good at language generation,  
they also enable a new type of machine learning, called {\em in-context learning} \citep{brown2020language}. In-context learning, also known as prompt-based learning, is an emergent ability that occurs in LLMs beyond a certain size (hundreds of
billions of parameters---less, with judicious prompting) that have been finetuned for conversational responses \citep{wei2022emergent}. In-context learning is inference-time, prompt-based, few-shot learning with instructions. As opposed to finetuning, model parameters are not adapted by in-context learning. 
%
%

Language generation tasks are  solved well by LLMs
with prompt-based learning. 
On the other hand, tasks that require reasoning, 
such as grade school math word
problems, are more difficult for LLMs \citep{cobbe2021training}.  Spurred-on by the impressive performance on language 
tasks,
much research has focused on understanding the reason for the poor performance of LLMs on 
reasoning tasks, and how it can be improved. 
Among this research, the Chain-of-thought paper
by \citet{wei2022chain} stands out. This work, and later work by \citet{kojima2022large}, showed that adding a simple instruction to the prompt, 
{\em Let's think step by step}, can provoke an LLM to perform the required intermediate reasoning
steps to answer difficult problems in a multi-step approach. 
%
Subsequently, performance on math word benchmarks  has increased markedly.
Much of the  performance increase of models 
such as OpenAI o1, o3 \citep{huang2024o1} and DeepSeek R1 \citep{shao2024deepseekmath,guo2025deepseek}
is attributed to multi-step reasoning methods as reviewed here, with reinforcement learning playing an increasingly important role, both for in-context learning and  finetuning \citep{xu2025towards,wu2024comparative,team2025kimi,chu2025sft}.  

The line of research into multi-step LLM-reasoning was started with grade school math word problems, with the GSM8K benchmark \citep{cobbe2021training}. Soon, other reasoning domains were included, such as reasoning about advanced math problems, computer code, robotic movement, and games. Many works have been published that build on Chain-of-thought  
\citep{chu2024navigate}. In this paper, we survey the literature using  a straightforward taxonomy. We discuss papers based on several reasoning benchmarks, as well as directly-related follow up work. 

Having started with basic math word problems, multi-step LLM reasoning approaches  now perform  logic reasoning, planning, combinatorial games, and robotic actions, amongst others. Some approaches do so by prompting the LLM to generate code that is then interpreted by external tools. Many studies in multi-step methods are using reinforcement learning, for finetuning,  in context reinforcement learning,  external optimization loops, and  self-reflection.
The main  contributions of this paper are: 
(1) we provide a survey of relevant approaches in multi-step reasoning with LLMs, 
(2)   
we propose a taxonomy based on the reasoning literature (step
  generation, step evaluation, and 
  control of reasoning steps), and (3) we formulate a research agenda for reasoning with LLMs.

This survey is organized as follows. Section~\ref{sec:background}
provides background information on the most relevant developments in LLMs,
including in-context learning and finetuning.  Of great importance are the benchmarks that are used in this
field (Section~\ref{sec:benchmarks}). Next, in Section~\ref{sec:taxonomy} we
provide a taxonomy of the field, where we discuss the main approaches in
detail. Then, in
Section~\ref{sec:discussion} we discuss our findings in a broader perspective. We also discuss
the relation between multi-step reasoning and work on self-reflection and
metacognition. This section concludes with an agenda for future
research.  Finally, Section~\ref{sec:conclusion} concludes the
survey. 

\section{Background, Scope and Selection of Papers}\label{sec:scoping}
Reasoning has a long and active history in  AI, in logical inference, and in other fields,  such as commonsense and analogical reasoning \citep{lewis2024evaluating,mitchell1990emergence}. Indeed, some of the early criticism on LLMs was that they could not reason,  that they made obvious and basic reasoning errors, showing a  lack of  understanding.  \citet{berglund2023reversal}  provide the example  of a model that is trained to report that ``Valentina Tereshkova was the first woman to travel to space,'' but is not able to answer the question, ``Who was the first woman to travel to space?'' pointing  to a lack of semantic understanding. Other work suggests that results are less generalizable and transferable than often assumed, showing  how base-10 arithmetic skills do not transfer to base-9 arithmetic problems \cite{wu2024reasoning}. 
More recently, LLMs are also shown to struggle with analogies  and analogical reasoning \citep{lewis2024evaluating,mitchell1990emergence} (see Sections~\ref{sec:self-reflection} and \ref{sec:metacog}). 
Different kinds of reasoning  play an important role in LLM research.

\subsection{Scope}
The question whether LLMs can reason prompted the development of the GSM8k benchmark \citep{cobbe2021training} in 2021. GSM8k is a benchmark of 8500 easy reasoning problems (of the type: ``Question: Natalia sold clips to 48 of her friends in April, and then she sold half as many clips in May. How many clips did Natalia sell altogether in April and May? Answer: 72''). This benchmark was specifically designed to test if LLMs can answer basic mathematical multi-step reasoning questions, or if they can not reason at all. Initially, performance was poor, but a year later \citet{wei2022chain} showed that  by providing multi-step examples in the prompt---a {\em chain of thought}---much better performance was possible. Many works followed, and reasoning in language models has become an active area of study. In this survey, we focus on prompt-based multi-step reasoning algorithms.
After starting with math word problems, work has more recently been extended to combinatorial games, puzzles, robotics, logic, and other fields of reasoning.
At the end of this survey, we  discuss connections to other fields, such as
self-reflection and in-context reinforcement learning.  

Before we discuss the works on reasoning, this section reviews 
background terminology on LLMs.  Our overview is
brief and we focus on the multi-step reasoning approaches that were inspired by Chain-of-thought. Excellent related surveys exist. For a longer overview of general LLM topics, see, for example, the surveys \citet{minaee2024large} and \citet{zhao2023survey}. For surveys that focus on the nature of reasoning and its definition, see \citep{yu2023natural,chu2023survey,huang2022towards}. For surveys on evaluating logical reasoning in LLMs, see \citep{mondorf2024beyond,liu2023evaluating,pan2023logic}. For a survey on reinforcement learning in reasoning, see \citep{xu2025towards}, for reasoning in language see \citep{yu2023natural}, and for a survey on Chain-of-thought itself, see \citep{chu2024navigate}.

The papers for this survey were selected as follows. 
We started by selecting papers on the ability to solve math word benchmarks (as a proxy for reasoning ability), that contained the search terms {\em reasoning} and {\em large language model} in their title or abstract, with a focus on papers that reference the Chain-of-thought paper. 
Although multi-step reasoning with LLMs initially was aimed at solving math world problems, 
it is now  wider, including benchmarks and approaches for computer code, game play, puzzles, robot movement, and webpage navigation (see Table~\ref{tab:geneval2}). 
We selected recent papers (two years prior to the writing of the survey) that show experimental results on selected benchmark datasets.

We focus on prompt-based, in-context learning, methods based on Chain-of-thought, that are used in reasoning LLMs such as OpenAI o1 and o3 \citep{xu2025towards,wu2024comparative,huang2024o1}. We  include a few papers that work with external algorithms,  finetuning, or supervised learning, that have contributed significantly to the in-context approaches.


\label{sec:background}
We discuss the generic training pipeline for LLMs,
we discuss how in-context learning works, and we discuss commonly used benchmarks. 
We start with the generic language model training pipeline.


\subsection{Language Model Training Pipeline}
LLMs are typically constructed in a sequence of
stages, from data preparation, through training, to inference. The
training pipeline for most LLMs is quite
elaborate. 
We will now list a brief pipeline of the most commonly used stages, based on the
survey by  \citet{minaee2024large}. 

In training an LLM, the first stage is to {\em acquire} a large, general, unlabeled, high-quality text corpus. Some considerations on the selection of the texts are discussed in  \citet{brown2020language}.
%
The next stage is {\em pretraining} the transformer model \citep{vaswani2017attention} on this large corpus. This stage yields a generative language
  model. Pretraining is done using self-supervised autoregressive training on the unlabeled dataset (text corpus).
Then the general model is 
{\em finetuned} to a specific (narrow) task, for example, question answering. This can be
  done using 
  supervised learning with a new labeled dataset consisting of prompts and answers (supervised
  finetuning, SFT), for example with low-rank optimization \citep{hu2021lora,wei2022emergent,minaee2024large}, or reinforcement learning \citep{sutton2018reinforcement,chu2025sft}. 
%
A specific form of finetuning is {\em instruction tuning}, to improve instruction following for a certain task. Instruction tuning is supervised by a labeled dataset of instruction prompts and corresponding outputs.
%
Depending on the purpose of the model the next step is {\em alignment} of the finetuned model with user expectations
(preference alignment). Alignment is usually performed to ensure that the model produces more ethically and socially acceptable answers, preventing, for example, hate speech. 
A machine learning
  method that is commonly used in this stage is Reinforcement
  Learning with Human Feedback  \citep{ouyang2022training},
  Direct Preference Optimization  \citep{rafailov2024direct}, or Reinforcement Learning with Verifiable Rewards \citep{lambert2024tulu}.
%
Optionally,  model training can be {\em optimized} to improve cost-effectiveness, for
  example, mixed precision training \citep{micikevicius2017mixed}, quantization \citep{jacob2018quantization}, or knowledge
  distillation \citep{xu2024survey,gu2023minillm}.

Once the model has been trained in the steps described above, it can be used further in the {\em inference} stage.  
Here, the model is used to provide an answer to the prompt. 
The inference-time stage is post-training, 
no model parameters are changed anymore
  \citep{dong2022survey,brown2020language}; {\em in-context learning}, or {\em prompt-learning}, takes place in this stage. 
  
In-context learning has a low barrier of entry. Since there is no need for expensive pretraining or finetuning, this form of advanced prompt engineering has become a popular training method.
  This is the stage on which
  most of the surveyed papers focus,
%
%
using prompts 
for the LLM to perform a complex multi-step reasoning task. The following section provides a brief introduction to in-context learning.

\subsection{In-Context Learning}

In large models, beyond hundreds of billions of parameters, a new kind of machine learning has emerged, that has been called {\em in-context learning} or  {\em prompt-learning} \citep{brown2020language}. It occurs not when the model's parameters are trained, but when the model is used, at inference time. Since no parameters are changed at this stage, it is not a {\em model training} stage; in-context-learning ``learns'' inside the context, or prompt, using information that is already encoded in the trained model parameters and the prompt, not by training the model anymore. In-context learning is often able to  give good results with few examples, so-called {\em few-shot} learning, learning from the few examples in the prompt in combination with the knowledge of the model. The large size of the model, containing rich and general knowledge, is enabling the
few-shot learning (see \citet{dong2022survey} for a survey).

In in-context learning, a prompt, consisting of a piece of
demonstration context, is concatenated with a query question, and is given
to the language decoder model, for text generation  \citep{liu2023pre}. For example,
when the task is emotion recognition in a social media post, ``I
missed the bus today,'' can be followed by ``I felt so [\_\_\_]'', and the model could answer with ``bad''.
Alternatively, for translation, we could follow ``I missed the bus today,'' by
``French: [\_\_\_]'' to request a translation \citep{liu2023pre}.
The prompt contains background information that is recognized by the
model, selecting the desired model context. In-context learning works when language models contain enough 
knowledge, allowing them to generalize on the (few) examples provided in the prompt.

Contexts that contain a few examples
are said to perform few-shot learning. Contexts that contain
only an instruction, with zero examples, are said to perform zero-shot learning.
%
%
%
%
In-context learning takes place at
inference time, after the computationally intensive training stages where parameters have been pretrained and finetuned, when the model is queried by the user to provide answers. No parameters are changed anymore with in-context 
learning. This is quite different from the common approach in
supervised deep learning---or self-supervised deep learning---where
large datasets are used during training to update model parameters
with backward propagation in lengthy and costly training epochs \citep{goodfellow2016deep}. Indeed, in-context learning takes place fully at inference time, no parameters are trained, instead,  {\em learning} now refers to adjusting the answers to the examples in the prompt and the internal knowledge acquired during training. Common approaches to few-shot learning, such as metalearning, do include training and finetuning of parameters to achieve generalization, and are computationally expensive (see, for example, \citep{finn2017model} or \citep{huisman2021survey,hospedales2021meta} for a survey). In-context learning, in comparison, is computationally cheap, and has become a popular research approach.

Prompts provide a user-friendly interface to LLMs. The success of in-context learning tends to be quite sensitive to the way in which a prompt is formulated; a
new field called {\em prompt engineering} has emerged to 
help human users to learn how to
make LLMs do what they want them to do \citep{radford2019language,wei2022emergent,giray2023prompt,sahoo2024systematic}. The current survey thus discusses advanced prompt engineering methods.


\subsection{Multi-step Reasoning Benchmarks}\label{sec:benchmarks}
Progress in artificial intelligence is measured by benchmarks. Benchmarks define the goal that
researchers aim to achieve in their experiments. In natural language processing, a wide array of benchmarks exist to measure progress, such as on question answering (for example, CommonsenseQA \citep{talmor2018commonsenseqa}), word prediction (for example, LAMBADA
\citep{paperno2016lambada}), translation (for example, WMT'22
\citep{kocmi2022findings}), language understanding (for example, GLUE \citep{wang2018glue,wang2019superglue}), and text
summarization (for example, Xsum \citep{narayan2018don}).

The field of LLMs is quite active. 
%
We will mention relevant benchmarks
for testing the multi-step reasoning abilities 
of LLMs. The research on reasoning with LLMs started with math word problem benchmarks.
The benchmark that is most frequently associated with multi-step reasoning with LLMs is the dataset of grade school
math word problems GSM8K \citep{cobbe2021training}.
GSM8K was created with the aim of providing high quality, high
diversity, moderate difficulty, problems and solutions in natural language. It consists of 8500 human-written math
problems. Language models struggled to achieve good performance on this dataset before Chain-of-thought was introduced.
An example of a math word task follows. {\em Problem:} Beth bakes 4 trays with 
two dozen batches of cookies in a week. If these cookies are shared amongst 16 people equally, how many cookies does
each person consume? {\em Answer:} $4\times 2\times 12/16=6$.

Other  benchmarks of similar math word problems are the SVAMP varying structures benchmarks \citep{patel2021nlp}, the
ASDiv dataset of diverse math problems \citep{miao2021diverse}, the AQuA dataset of algebraic
word problems \citep{ling2017program}, and the MAWPS benchmark
\citep{koncel2016mawps}. Table~\ref{tab:mwpbench} summarizes the accuracy of Chain-of-thought on these basic math word problems, against the baseline of  GPT-3 175B as LLM \citep{wei2022chain}, as percentage of benchmark questions answered correctly. 
\begin{table}
\begin{center}
    \begin{tabular}{ccc}
    Benchmark & GPT-3 175B & Chain-of-thought \citep{wei2022chain} \\ \hline
    GSM8K & 15.6 & 46.9\\
    ASDiv & 70.3 & 71.3\\
    MAWPS & 72.7 & 87.1\\
    SVAMP & 65.7 &68.9\\
    AQuA  & 24.8 & 35.8\\
    \hline
    \end{tabular}
    \caption{Accuracy of GPT-3 and Chain-of-thought on popular Math Word Problems benchmarks \citep{wei2022chain}}\label{tab:mwpbench}
\end{center}
\end{table}
We see that Chain-of-thought performs well against the baseline of GPT-3 on some benchmarks, but there is certainly room for further improvement on others. 

In addition to the initial set of math word benchmarks, further reasoning approaches have been introduced that test performance in other fields of reasoning. Benchmarks have been developed on Advanced mathematical questions \citep{codeforcesamerican},
Computer code comprehension (Human evaluation, Spider \citep{yu2018spider}, Transcoder \citep{roziere2020unsupervised}), Robotic movement (Alfworld \citep{shridhar2020alfworld}, Kitchen \citep{ahn2022can}), Puzzle solving (Game24 \citep{yao2024tree}), Creative writing \citep{yao2024tree}), Gaming (Checkmate problems\citep{yang2024buffer}, MineCraft \cite{fan2022minedojo}), and Webpage navigation (WebShop \citep{yao2022webshop}). These benchmarks are used by other approaches in our survey, as we will see in more detail in Section~\ref{sec:taxonomy}. 

The scope of this survey is limited to multi-step reasoning approaches. Other studies have published more challenging benchmarks, to study the performance of Chain-of-thought and self-reflection in LLMs in puzzles and (logic) games \citep{ruoss2024lmact,paglieri2024balrog,wen2025thinkpatterns}. 

In this survey we will mention benchmark results as reported by the surveyed papers, for the sake of concreteness. However, it  can be difficult to directly compare different benchmarking results, since there may be differences in the way the experiments were conducted. \citet{kamoi2024can} provide a critical analysis of benchmarking problems in LLM reasoning. Despite these challenges, general conclusions can be drawn. Section~\ref{sec:guidance} provides guidance on how to choose reasoning approaches for different problem types.

\section{Step Generation, Evaluation and
  Control}\label{sec:taxonomy}
This survey examines how LLMs based on the transformer architecture 
can be prompted to solve multi-step reasoning 
tasks. The Chain-of-thought paper shows how a simple command
can prompt an LLM to perform reasoning steps, yielding much better
performance in math word problems. Much research  has further explored this approach, trying to 
build stronger general problem solvers for other types of reasoning problems.

\label{sec:reasoning}
A typical approach to solve a complex problem is to subdivide it into
smaller steps and to solve those. This approach is related to classical divide
and conquer \citep{bellman1966dynamic}. It consists of three stages. New steps are (1) {\em generated}, (2) {\em evaluated}, and the search of the generated steps is (3) {\em controlled} in some way. 
The in-context reasoning approaches that we survey follow a general three-stage pipeline \citep{madaan2023self}:
\begin{enumerate}
\item {\bf Generate}: prompt the model to  generate reasoning steps,
\item {\bf Evaluate}: prompt the model to evaluate the generated steps,
\item {\bf Control}: prompt the model to control the number of steps that are
  generated and how deep ahead the reasoning process will look.
\end{enumerate}
This three-stage pipeline is the basis of our taxonomy. 
%
%
We will now discuss the three stages more deeply;  for ease of reference they are numbered according to the Subsection in which they are described in more detail  (3.1, 3.2, 3.3). Please also refer to Figure~\ref{fig:dia1}, and to Table~\ref{tab:geneval2}, for a diagram of the categories and subcategories of
different  approaches for the generation, evaluation, and
control of reasoning steps.\footnote{We show the approaches in the Figure in their main category only. Some approaches show innovations in two categories, and are shown twice.  (Since all approaches have a generation, an evaluation, and a control aspect, all could in principle  occur three times, and all three columns can be found in Table~\ref{tab:geneval2}).} 

\begin{figure}
  \begin{center}
  \usetikzlibrary{positioning, arrows.meta, calc}

\tikzset{
  cat/.style={
    rectangle, draw=gray!70, fill=gray!10,
    rounded corners=3pt, inner sep=6pt,
    font=\sffamily\bfseries, align=center
  },
  sub/.style={
    rectangle, draw=gray!70, fill=white,
    rounded corners=3pt, inner sep=5pt,
    font=\sffamily\normalsize,
    text width=3cm, align=left,
    align=center
  },
  refs/.style={
    rectangle, draw=gray!70, fill=gray!05,
    rounded corners=3pt, inner sep=5pt,
    font=\sffamily\footnotesize,
    text width=5cm, align=left,
    minimum height=1cm, anchor=west
  },
  arr/.style={-{Stealth[length=3mm]}, very thick, gray!70},
  narr/.style={-, very thick, gray!70}   
}

\begin{tikzpicture}

\coordinate (root) at (0,0);
  \node[cat, right=0cm of root, yshift=+5cm] (gen)  {1.\;Generation};
  \node[cat, right=0.8cm of root]              (eval) {2.\;Evaluation};
  \node[cat, right=1.6cm of root, yshift=-7cm] (ctrl) {3.\;Control};

\draw[arr] (gen.west) |- (eval.west);
\draw[arr] (eval.west) |- (ctrl.west);

  \coordinate (refcol) at ([xshift=10cm]root);

  \node[sub, right=2cm of gen, yshift=+1.6cm] (hwp) {Hand-written prompt};
  \node[sub, right=2cm of gen]                (mgp) {Model-generated prompt};
  \node[sub, right=2cm of gen, yshift=-1.3cm] (pek) {Prompt using\\external knowledge};

  \coordinate (tmpH) at (hwp.west -| gen.east);
  \draw[narr] (gen.east) -- (tmpH);
  \draw[arr]  (tmpH)       -- (hwp.west);

  \draw[arr]  (gen.east) -- (mgp.west);

  \coordinate (tmpP) at (pek.west -| gen.east);
  \draw[narr] (gen.east) -- (tmpP);
  \draw[arr]  (tmpP)       -- (pek.west);

  \node[refs] (hwprefs) at (refcol|-hwp) {
    Scratchpad; \citet{nye2021show} [supervised]\\
    Chain-of-Thought; \citet{wei2022chain}\\
    ZS-CoT; \citet{kojima2022large}
  };
  \draw[arr] (hwp.east) -- (hwprefs.west);

  \node[refs] (mgprefs) at (refcol|-mgp) {
    Auto-CoT; \citet{zhang2022automatic}] \\
    Complexity; \citet{fu2022complexity} \\
    Buffer-of-Thoughts; \citet{yang2024buffer}
  };
  \draw[arr] (mgp.east) -- (mgprefs.west);

  \node[refs] (pekrefs) at (refcol|-pek) {
    Self-Ask; \citet{press2022measuring}
  };
  \draw[arr] (pek.east) -- (pekrefs.west);

  \node[sub, right=2cm of eval, yshift=+2cm] (sa)  {Self-Assessment};
  \node[sub, right=2cm of eval]                (tbe) {Tool-based\\Evaluation};
  \node[sub, right=2cm of eval, yshift=-3cm] (emv) {External Model\\Validation};

  \coordinate (tmpSA) at (sa.west -| eval.east);
  \draw[narr] (eval.east) -- (tmpSA);
  \draw[arr]  (tmpSA)       -- (sa.west);

  \draw[arr]  (eval.east) -- (tbe.west);

  \coordinate (tmpEM) at (emv.west -| eval.east);
  \draw[narr] (eval.east) -- (tmpEM);
  \draw[arr]  (tmpEM)       -- (emv.west);

  \node[refs] (sarefs) at (refcol|-sa) {
    Self-Verification; \citet{weng2022large} \\
    Self-Consistency; \citet{wang2022self}
  };
  \draw[arr] (sa.east) -- (sarefs.west);

  \node[refs] (tberefs) at (refcol|-tbe) {
    Codex; \citet{chen2021evaluating} \\
    Self-Debugging; \citet{chen2023teaching} \\
    Fun-Search; \citet{romera2024mathematical} \\
    LLaMea; \citet{van2024llamea} \\
    MathPrompter; \citet{imani2023mathprompter} \\
    Program-of-Thoughts; \citet{chen2022program} \\
    Program-Aided-Language; \citet{gao2023pal}
  };
  \draw[arr] (tbe.east) -- (tberefs.west);

  \node[refs] (emvrefs) at (refcol|-emv) {
    \textbf{Finetuning:}\\
      \hspace*{0.6em}Refiner; \citet{paul2023refiner} \\
      \hspace*{0.6em}Self-Corrector; \citet{welleck2022generating}\\
      \hspace*{0.6em}Self-Improvement; \citet{huang2022large}\\[2pt]
      \textbf{Robot:}\\
      \hspace*{0.6em}SayCan; \citet{ahn2022can}\\
      \hspace*{0.6em}Inner Monologue; \citet{huang2022inner}\\[2pt]
      \textbf{Data Augmentation:}\\
    Self-Taught Reasoner; \citet{zelikman2022star}
  };
  \draw[arr] (emv.east) -- (emvrefs.west);

  \node[sub, right=2cm of ctrl, yshift=+1.7cm] (gs) {Greedy Selection};
  \node[sub, right=2cm of ctrl]                (es) {Ensemble Strategy};
  \node[sub, right=2cm of ctrl, yshift=-2.7cm] (rl) {Reinforcement Learning};

  \coordinate (tmpGS) at (gs.west -| ctrl.east);
  \draw[narr] (ctrl.east) -- (tmpGS);
  \draw[arr]  (tmpGS)       -- (gs.west);

  \draw[arr]  (ctrl.east) -- (es.west);

  \coordinate (tmpRL) at (rl.west -| ctrl.east);
  \draw[narr] (ctrl.east) -- (tmpRL);
  \draw[arr]  (tmpRL)      -- (rl.west);

  \node[refs] (gsrefs) at (refcol|-gs) {
    Complexity (Fu et al.\,2022)\\
    Least-to-Most (Zhou et al.\,2022)
  };
  \draw[arr] (gs.east) -- (gsrefs.west);

  \node[refs] (esrefs) at (refcol|-es) {
    Self-Consistency; \citet{wang2022self}\\
    Self-Verification; \citet{weng2022large}\\
    MathPrompter; \citet{imani2023mathprompter} \\
    Program-Aided-Language; \citet{gao2023pal}
  };
  \draw[arr] (es.east) -- (esrefs.west);

  \node[refs] (rlrefs) at (refcol|-rl) {
    Progressive-Hint; \citet{zheng2023progressive}\\
    Self-Refine; \citet{madaan2023self}\\
    Tree-of-Thoughts; \citet{yao2024tree}\\
    Buffer-of-Thoughts; \citet{yang2024buffer} \\
    Algorithm-of-thoughts; \citet{sel2023algorithm}
    
    Beam-Search; \citet{xie2024self} \\
    ReAct; \citet{yao2022react}\\
    Reflexion; \citet{shinn2024reflexion} \\
    Voyager; \citet{wang2023voyager}
  };
  \draw[arr] (rl.east) -- (rlrefs.west);

\end{tikzpicture}
    \caption{Taxonomy of LLM-Reasoning Approaches: Prompt Generation, Evaluation, and Control}\label{fig:dia1}
  \end{center}
\end{figure}

\paragraph{(3.1) Generation}
The first stage is to create a prompt that instructs the
LLM to generate reasoning steps. The problem
must be split into substeps. This can be achieved with  a problem-specific prompt that contains elements of
the problem, such as: ``First calculate how many marbles Mary had originally, then how many her friend has, and finally how many they have together.'' 
In general, it is possible to prompt an LLM to fill in the blanks in a step-by-step fashion. In
the papers that we  discuss, there are three
main approaches for generating the step-by-step prompt,  numbered with the Subsection in which the approaches are described. First, the prompt may be
handcrafted for the problem by the researchers: {\em (3.1.1) hand-written prompt}. Second, the prompt or prompts may come from a source that is 
external to the model, such as another model or  dataset: {\em (3.1.2) external knowledge-based prompt}. Third,   
the model itself can be prompted to generate a (series of) prompt(s) to analyze the problem  {\em (3.1.3) model-generated prompt}. As we will see, all
three approaches have their advantages and disadvantages.

Generating the subproblem-steps is the first stage that is necessary for
in-context learning to perform reasoning. Each paper in our survey
performs at least this stage of the reasoning pipeline. In some of the early 
papers (around 2022) it is the only stage of the pipeline that is performed.

\paragraph{(3.2) Evaluation}
After the prompt has been generated and the model has answered it, the
next stage is to evaluate the quality of this answer. Such evaluation is often necessary to improve the answer and perform well on the benchmark. Again, we see
three main approaches for substep evaluation. First, the
steps may be evaluated by the model itself: {\em (3.2.1) self-assessment}. Second, an external program can be used to evaluate the steps. For example,
when the steps are expressed as computer code, an external interpreter or
compiler can be used to check the validity of the outcome: {\em (3.2.2) tool-based evaluation}. Finally,  an external model
can be used, LLM or otherwise. For example, in robotics, an external physics model can
determine if certain actions are physically possible: {\em (3.2.3) external model validation}.

\paragraph{(3.3) Control}
The third stage is  control. A reasoning process that consists of multiple steps is  a sequential decision process \citep{littman1996algorithms}. When a single chain of reasoning steps is generated, the control flow of
the reasoning process is simple: greedily evaluate the first step and
then the next one, if present.  The control flow of the reasoning process may also be  more intricate. Some reasoning problems can be divided into multiple subproblems.  To
execute, evaluate and combine the results of all substeps, a
separate controller may be needed. This controller can be a prompt or an external
algorithm.    

Again, we distinguish three approaches.
Most papers use  a {\em (3.3.1) greedy selection} approach: a single prompt with a single chain of steps is generated, and
these steps are directly executed and followed.  The second
approach  is to generate an {\em (3.3.2) ensemble strategy} of reasoning steps, evaluate
them, combine the individual results, and present them as the result
of the ensemble.
Finally,  a full tree-search or a {\em (3.3.3) reinforcement learning} (RL)
algorithm can be used as scaffolding \citep{plaat2020learning}. In this case, when a step is followed and evaluated, the LLM 
can roll back and try a different reasoning step. Going further, a full reinforcement learning approach
can be used \citep{sutton2018reinforcement,plaat2022deep} to find an optimal policy for the sequential decision process. In general a  Markov Decision Process of state, action, transition, and 
reward function can be specified, and step control can become a  process where prompts are generated dynamically under the control of an external RL algorithm, or as in-context reinforcement learning (ICRL) \citep{lee2023supervised,moeini2025survey}. 

\begin{table}
\begin{footnotesize}
  \caption{Taxonomy of approaches: Generation, Evaluation, and Control. Reported benchmark results: '=' is absolute score, '+' is improvement to a baseline}\label{tab:geneval2}
  \begin{adjustbox}{width=\textwidth}
  \begin{tabular}{llcccl}\hline\hline
    {\bf Approach} & {\bf Domain} & {\bf (3.1) Step generation} & {\bf (3.2) Step evaluation} & {\bf (3.3) Step control} & {\bf Result}\\ \hline

    Scratchpad   \citep{nye2021show} &math word&{\bf hand-wr/superv} & -  & greedy/1prompt & \scriptsize PolyEval +19\%, Python +21\% \\
    Chain-of-thought \citep{wei2022chain} &math word&{\bf hand-written}   & -  & greedy/1prompt &\scriptsize GSM8K +39\%, SVAMP +10\%, ASDiv +2\%, AQuA +11\%,\\
                                          &         &                     &    &                &\scriptsize  MAWPS +14\%, CSQA +1.8\%, StrategyQA +0.2\% \\
    ZS-CoT    \citep{kojima2022large} &math word&{\bf hand-written}   & -  & greedy/1prompt & \scriptsize MultiArith =89\%, GSM8K =70\% \\
    Auto-CoT  \citep{zhang2022automatic} &math word&{\bf model-generated} & -  & clustering & \scriptsize MultiArith +0.3\%, GSM8K +1\%, AddSub +3.5\%,\\
    &         &                      &    &             & \scriptsize   AQuA +0.7\%, SingleEq +0.4\%, SVAMP +0.6\%,   \\
    &&&&& \scriptsize CSQA +1\%, StrategyQA +0\%, Letter +0.7\%, Coin +2.7\% \\ 
    Complexity   \citep{fu2022complexity} &math word& {\bf hand-written }&  self-consistency &  greedy/1prompt & \scriptsize GSM8K +7\%, MultiArith +3\%, 
    Penguins +3\% \\
    Self-ask   \citep{press2022measuring} &math word&{\bf ext knowledge} & LLM&  multi-hop q.& \scriptsize Bamboogle =60\%, 2Wiki =40\%, Musique =15\% \\ \hline
    Self-verification \citep{weng2022large}   &math word& hand-written   & {\bf back-verify} & ensemble & \scriptsize GSM8K +4\%, SingleEq +2\%, AddSub +4\%, \\
                                              &         &                 &                   &          & \scriptsize MultiArith +3\%, AQuA +3\%, SVAMP +1\% \\
    Self-consistency \citep{wang2022self}    &math word& hand-written   & {\bf majority} & ensemble & \scriptsize GSM8K +18\%, SVAMP +11\%, AQuA +12\%, \\
                                             &&&&& \scriptsize StrategyQA +6\%, ARC-c +4\% \\ 
    Codex   \citep{chen2021evaluating}       &  code    &    -          & {\bf tool-based }& - & \scriptsize HumanEval =70\% \\
    Self-debugging \citep{chen2023teaching}  &code     & hand-written  & {\bf tool-based }& greedy & \scriptsize Spider +9\%, MBPP +12\%, TransCoder +12\% \\
    Fun-search \citep{romera2024mathematical}  & code   &  hand-written    & {\bf tool-based           }  & evolutionary alg &\scriptsize cap set 8 =512 \\
    LLaMEa \citep{van2024llamea}      & code   & hand-written          & {\bf tool-based }& evolutionary alg& \scriptsize BBOB +11\% \\
    MathPrompter  \citep{imani2023mathprompter} &math     &hand-written   & {\bf tool-based }& ensemble & \scriptsize MultiArith =92\% \\
    Program-of-thoughts \citep{chen2022program} & math word&hand-wr, Codex   & {\bf Python+Consist.}  &       split reason/cmput & \scriptsize GSM8K =71\%, SVAMP =85\%, ASDIV =85\%, \\
    &&&&& \scriptsize AddSub =92\%, MultiArith = 99\% \\
    Program-aided-lang \citep{gao2023pal}& math word&hand-wr, Codex & {\bf  NLP/Python }&   ensemble & \scriptsize GSM8K =72\%, SVAMP =79\%, ASDIV =79\%,  \\
    &&&&& \scriptsize SingleEQ =96\%, SingleOP =94\%, AddSub = 92\%, \\
    &&&&& \scriptsize MultiArith = 99\%, Penguins = 93\% \\
    Refiner   \citep{paul2023refiner}        &math word &finetune            & {\bf critic model  }  & gen/crit feedback & \scriptsize SVAMP =72\%, GSM8K =78\% \\
    Self-correction \citep{welleck2022generating} &math word     &finetune    & {\bf corrector model  }   & gen/corr feedback & \scriptsize MathProgSynth =24\%, LexConstrGen =98\%,\\
    &&&&& \scriptsize ToxicityControl =0.0\% \\
    Self-improvement \citep{huang2022large} &math word          &finetune    & {\bf self-assessment }   & CoT/consistency & \scriptsize GSM8K =82\%, DROP =83\%, ARC-c =90\%, \\
    &&&&& \scriptsize OpenBookQA =94\%, ANLI-A3 =68\% \\ 
    Say-can     \citep{ahn2022can}          & robot  &model-generated        & {\bf external model }  & greedy & K\scriptsize itchen =31\% \\
    Inner-monologue  \citep{huang2022inner}     &robot    &hand-written      & {\bf various       }    & greedy & \scriptsize TableTop =90\%, Kitchen =60\% \\ 
    Self-taught-reasoner \citep{zelikman2022star} &math word&finetune        & {\bf  augmentation}& greedy/feedback&\scriptsize CommonsenseQA =72\% \\ \hline
    Least-to-most  \citep{zhou2022least} &math word& hand-written   &  self-assessment    & {\bf curriculum} & \scriptsize SCAN =99\% \\
    Progressive-hint \citep{zheng2023progressive}    &math word& model-generated &  self-assessment    & {\bf stable prompt} &\scriptsize AddSub +2\%, MultiArith +0\%, SingleEQ +2\%,\\
    &&&&& \scriptsize SVAMP +3\%, GSM8K +5\%, AQuA +1\% \\ 
    Self-refine   \citep{madaan2023self}       &math word&model-generated &  self-assessment    &  {\bf greedy/feedback}& \scriptsize Sentiment +32\%, Dialog +49\%, CodeOptim +8\%,\\
    &&&&& \scriptsize  CodeRead +28\%, MathReason +0\%, \\
    &&&&& \scriptsize AcronymGen +25\%, ConstrainedGen +30\% \\
    Tree-of-thoughts \citep{yao2024tree}    &puzzles  &model-generated & self-assessment   & {\bf BFS/DFS} &\scriptsize Game24 =74\%, CreativeWriting , Crossword \\
    Buffer-of-thoughts \citep{yang2024buffer}& math word & thought template         &  self-assessment          &   {\bf buffer manager} &\scriptsize Game24 +11\%, GeoShapes +20\%, Checkmate +51\% \\
    Algorithm-of-thoughts \cite{sel2023algorithm}& puzzles & model-generated & self-assessment & {\bf in-context RL} & \scriptsize GSM8K =89\%, StrategyQA =84\%, Crossword \\
    Beam-search  \citep{xie2024self}        &  math word       &model-generated & self-assessment    & {\bf Beam Search} & \scriptsize GSM8K +6\%, AQuA +9\%, StrategyQA +5\% \\
    ReAct       \citep{yao2022react}          &action   & external knowledge & self-assessment    & {\bf reinf learning}& \scriptsize ALFWorld =34\%, WebShop =10\% \\
    Reflexion  \citep{shinn2024reflexion}          &decision &model-generated & ext model   & {\bf reinf learning} & \scriptsize HumanEval =91\% \\
    Voyager  \citep{wang2023voyager}             &Minecraft&model-generated & Minecraft   & {\bf reinf learning} & \scriptsize 15 x faster \\ \hline
  \end{tabular}
  \end{adjustbox}
  \end{footnotesize}
\end{table}


\

\paragraph{Taxonomy Table}
Table~\ref{tab:geneval2} lists the main papers of this survey. We show the domain they work on,  the type of
prompt generation, the evaluation of the result, and the control method. 
These three categories of approaches---indicated by their Sections (3.1) generation, (3.2) evaluation, (3.3) control---are shown in the table as groups divided by horizontal lines. 
The first group in the Table, from Scratchpad to Self-ask, focuses on creating a prompt that {\em generates} the reasoning steps. The entries in the cells of this column are shown in bold, highlighting the focus of the approaches. The approaches in this group are  the start of the field of LLM-reasoning. The Chain-of-thought approach is especially an inspiration for many works. The prompts are often written manually by the researchers for each problem; the steps are encoded in one prompt, and step control is greedy. There is no specific evaluation of the steps, other than comparing results to the benchmark. The Scratchpad approach is special in that it uses supervised learning, not prompt-learning; the work showed that LLMs can generate internal reasoning steps by supervised learning, paving the way for in-context  works. 

The second group, from Self-verification to Self-taught-reasoner, focuses on {\em evaluation} of the reasoning steps in the prompt. This column is shown in bold in the table. The approaches in this group aim to improve the Chain-of-thought results by reducing error accumulation that occurs when multiple steps are taken in a reasoning chain.  A variety of step control methods is used by these approaches, which is discussed in more detail later.
Note that not all approaches use natural language problems. For example, the subgroup of Codex to Program-aided-language focuses on formal languages. They generate code or math equations, typically in Python, to formalize the steps of the reasoning problem, or as result of the task. LLMs that are trained on computer code achieve good performance at code generation,  \citet{chen2021evaluating} report up to 77\% accuracy on the HumanEval dataset. \citet{gao2023pal} leverage this ability by translating problems into code, achieving between 61\% and 99\% accuracy on various mathematics tasks. The use of code also allows the approaches to call external programs such as interpreters and debuggers to evaluate the correctness of the reasoning steps that are generated. 

There is also a special subgroup, Refiner to Self-improvement, that uses finetuning in addition to prompt learning.  Here, new data is generated based on reasoning exemplars, which is then used to further train the model. The extra data is often generated as a separate dataset, sometimes called {\em critic} or {\em corrector}.

There are two approaches, Say-can and Inner-monologue, whose application domain is control of robot movement. Robotic movement is constrained by the laws of physics (both in the body of the robot  as in aspects of its environment). The laws of physics are learned and used to ground the reasoning steps in reality (to reduce hallucination).

The third group, Least-to-most to Voyager, addresses step {\em control} (shown in bold in this column). Whereas in the previous approaches the reasoning steps are written in a single, static, prompt, these approaches generate the steps in multiple, dynamic, prompts. This allows control of the space of reasoning steps.  Various search control approaches are used, all in the form of an external algorithm that performs calls to the LLM with different prompts. The  control methods range from simple greedy and depth-first search to elaborate beam search and reinforcement learning schemes. 

The last column of the Table summarizes reported benchmark results. The '=' symbol indicates absolute scores on the benchmarks, while '+' indicates relative improvement in percentage points over a baseline LLM, typically GPT-3.5. Results vary strongly, both between approaches and within a single approach between benchmarks. Also, different LLMs were used, from early stage to more mature models, open and commercial, and the baselines differ. For some benchmarks, such as the Creative Writing benchmark in Tree-of-thoughts, results are best reported qualitatively. The source papers provide more measurement details, Section~\ref{sec:guidance} discusses when to which approach to use for different applications.

In conclusion, we see a diverse array of methods that often achieve high performance in reasoning on their respective domains. To better understand the approaches, we discuss them in more detail, starting with the generation of steps.

\subsection{Generation of Steps}

\begin{figure}
  \begin{center}
    \includegraphics[width=7cm]{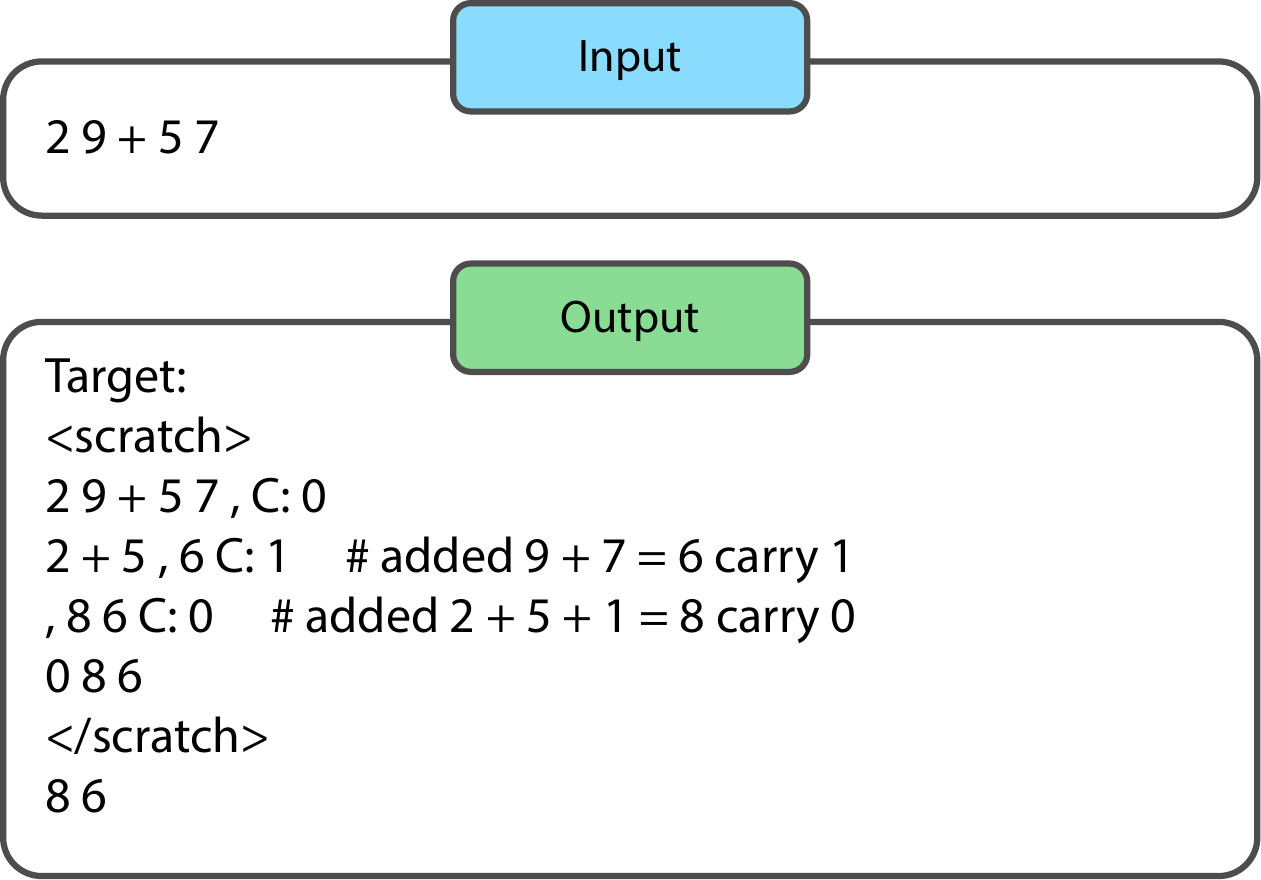}
    \caption{Example of input and target for supervised learning on a {\em long addition} problem of adding two numbers. The carry is
      recorded in the C: digit. Comments (after \#) are not part of the learning
      target (adapted from \citep{nye2021show})}\label{fig:scratchpad2}
  \end{center}
\end{figure}

Originally, LLMs performed poorly on math word problems such as GSM8K \citep{cobbe2021training}. Different approaches were tried unsuccessfully,  for example  scaling up the
size of the LLM~\citep{rae2021scaling}. 
%
%
%
The LLM architecture,
based on transformers, is designed to produce a single token. When we
prompt such an architecture to produce an answer, it does so. What we should do instead, is to prompt it to follow intermediate steps, answer those, and thus work towards the final answer, just as a student is taught to break down a complex problem into smaller steps.  We
should guide the model to explicitly produce intermediate steps, and combine the intermediate results.  
This idea was used by \citet{nye2021show} in Scratchpads, a transformer model that performs multi-step computations by asking it
to emit intermediate computation steps into a {\em scratchpad}.
They train the model by supervised learning (not prompt-based
in-context learning).
Figure 
\ref{fig:scratchpad2}  shows an  example.
On experiments with addition, polynomial evaluation, and
Python code execution, versions that produced the intermediate steps on a scratchpad performed considerably better than versions that did not, going from 35\% to 95\%, from 32\% to 51\%, and from 30\% to 42\% accuracy, respectively.


If supervised learning can produce intermediate steps, would prompt learning be able to do so, too?

\subsubsection{Hand-written Prompt}

\begin{figure}
  \begin{center}
    \includegraphics[width=\textwidth]{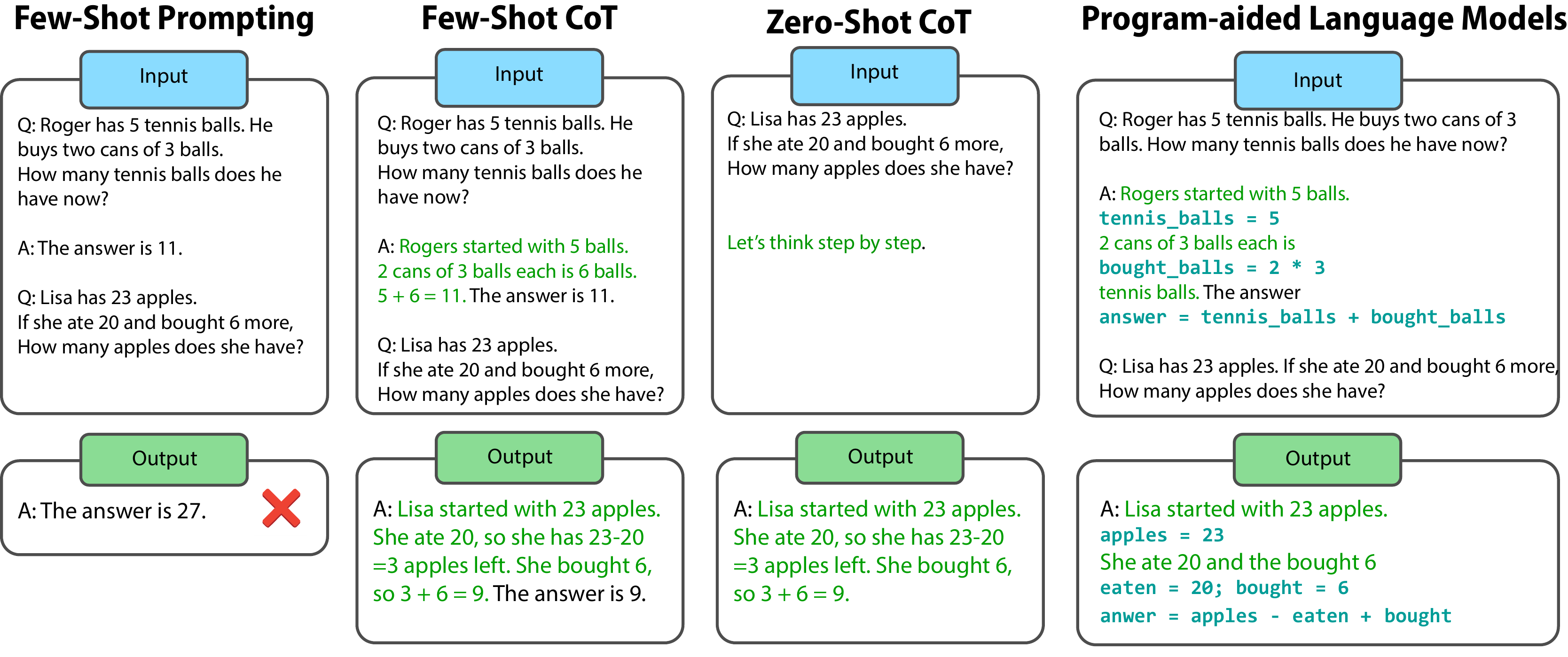}
    \caption{Different chain-of-though (CoT) prompting techniques. At the top the prompts, at the bottom the answers. When shown the longer example prompt,
      the LLM follows the longer example when answering the question  (Few-Shot CoT \citep{wei2022chain}). Without example answer and using \textit{Let's think step by step} results in similar answers (Zero-Shot CoT \citep{kojima2022large}). With Program-aided-language models \citep{gao2023pal} similar reasoning can be achieved. }\label{fig:cot} \label{fig:zero} \label{fig:pal} 
  \end{center}
\end{figure}

This question was studied by \citet{wei2022chain}, amongst others.
A basic way to instruct an LLM to generate steps by prompt-learning is to manually write a prompt for the large
language model to follow the reasoning steps.
%
%
%
When the LLM is prompted to rephrase information from the question as
intermediate reasoning steps in its answer, the LLM performed
much better than when it was prompted to answer a math problem directly, without
reproducing  information from the question in its answer in multiple steps. The example from the
Chain-of-thought paper is shown in Figure~\ref{fig:cot}.  Table~\ref{tab:mwpbench} shows that the largest accuracy increase is on GSM8K, from 16\% to 47\%. 

%
The idea that an LLM can be made to follow step-by-step instructions, and the performance improvement by
Chain-of-thought have caused much excitement and have opened up further research on reasoning with LLMs. 
%
In the
original  paper  the prompts were handwritten by
the researchers for the individual types of problems, and evaluations are conducted with  benchmarks  (not by an LLM).\footnote{The Chain-of-thought idea is about prompt generation, not
about the evaluation or the search control of the reasoning
steps. Hence, in  Table~\ref{tab:geneval2} Chain-of-thought is labeled as {\em greedy}
without an evaluation.} In a later work the prompts were generated automatically by
the LLM \citep{zhang2022automatic}, and evaluated.
%


\citet{kojima2022large} go a step further.
They show that the  addition of a single text to the prompt ({\em Let's think step by step})
significantly improves performance. Since this text does not contain problem-related elements, it is as a form of zero-shot learning.
Figure~\ref{fig:zero} (third column) compares the approaches.
Experiments further show that with this  addition to the prompt significant performance gains
are also achieved on other reasoning benchmarks, including  arithmetic,  symbolic, and  logical reasoning (achieving 70\% accuracy on GSM8K/PaLM  when Self-consistency is also included).

The Chain-of-thought idea itself is inspired by earlier work where natural language
steps are generated for arithmetic reasoning
\citep{ling2017program,cobbe2021training}, and the use of formal languages for reasoning~\citep{roy2016solving,chiang2018semantically,amini2019mathqa,chen2019neural}.

\subsubsection{Prompt using External Knowledge}
Chain-of-thought 
prompts are written manually, by the researchers, an approach that does not scale.
%
%
We can  use external information about the problem to improve the prompt.
\citet{press2022measuring} study how subproblems are related to the main problem, which they call {\em compositional reasoning}.
They study how often a
model is able to answer the subproblems, but
not the overall problem. This difference is called the compositionality gap. They find that in GPT-3, as model size increases,  
the single-hop question-answering performance improves faster than the
multi-hop performance: while more powerful
models memorize and recall more factual knowledge,  no
improvement in their compositional reasoning occurs. The
ability to reason does not depend on the size of the model.


Subsequently, a method called {\em Self-ask} is proposed, that asks
elicitive follow-up questions (like Chain-of-thought, but with the {\em follow up:} prompt), that the model then answers. 
Self-ask can also use an external  search engine to answer intermediate prompts, instead of the model. 
The initial subquestion is fed into the search engine, and  the answer is processed by the model, which generates another subquestion, and so on, until it produces the final answer.
Self-ask was tested on three  benchmarks that were specifically designed for
multi-hop questions. 
Although it performs only a few percentage points better than vanilla Chain-of-thought, it showed how external knowledge can be used in a reasoning setting.

\subsubsection{Model-Generated Prompt}
In addition to manually writing prompts or using external information, we can also let the
LLM itself study the problem to write the best reasoning-prompt.
An example of such self-improvement is Auto-chain-of-thought
\citep{zhang2022automatic}. This approach builds
on the observation by \citet{kojima2022large} that large language
models are zero-shot reasoners. First, Auto-chain generates specific questions for a
given dataset and partitions them into  clusters. Then an external algorithm uses the model to generate examples that are sampled for diversity. The constructed demonstrations augment the in-context prompt.
This approach also performed a few percentage points better than hand-written Chain-of-thought prompts, on ten benchmarks, using GPT-3 (see Table~\ref{tab:geneval2}).

\citet{fu2022complexity} introduce Complexity-based
prompting. Inspired by Chain-of-thought and Self-consistency, 
their work specifically studies the impact of the complexity  of the
reasoning chain (the number of steps), and introduces a related reasoning approach (Complexity-based prompting).
They find that prompts with the largest
complexity  perform best, and also that  answers with the highest complexity are the best.  Complexity-based prompting
achieves somewhat higher performance on three math reasoning benchmarks: GSM8K improves 7 points, MathQA 6 points, and the Penguins benchmark from Big Bench Hard improves 3 percentage points. 

We see that the initial approaches showed larger  improvements than the later approaches. It is time to look at another category of approaches, that focus on the evaluation of reasoning steps.


\subsection{Evaluation of Steps}

After discussing prompts for the generation of reasoning steps, the next stage in the generation/evaluation/control
pipeline  is {\em evaluation} of the results of the steps. This stage focuses on reducing error accumulation of multi-step reasoning chains.
We will start with approaches where the same model performs step-generation and step-evaluation.

\subsubsection{Self-Assessment}

When LLMs are prompted to perform reasoning steps, they 
perform a sequence of steps and predict multiple tokens. Performing a
sequence of steps makes them  
sensitive to mistakes and vulnerable to error
accumulation  (logical, factual, ethical, or otherwise) \citep{weng2022large,xiao2023survey}. Several methods
have been developed to prevent error
accumulation. One approach is to create a new model to separately
evaluate the results. \citet{shen2021generate} and 
\citet{li2022advance} train an external verifier to check  results.
In contrast, \citet{weng2022large} propose an automated approach using evaluation by
the same LLM, called Self-verification. They note that human reasoning also suffers from the 
problem of accumulating errors, and that in human reasoning we frequently revisit our thought process to verify the accuracy of our reasoning steps. 
The LLM is prompted to use
the conclusion of the Chain-of-thought reasoning chain as a 
condition for solving the original problem and then compare
the answer, going back to the original question. The LLM is given variations of its own conclusion and is instructed to choose the one with the
highest similarity to the original question. (Note that there can be feedback issues using an LLM to evaluate itself, for a discussion see \citet{zheng2024judging}.)
Experiments are reported on GPT-3 \citep{chen2021evaluating} and on
Instruct-GPT \citep{ouyang2022training}. The accuracy of
Chain-of-thought was improved by a few percentage points on arithmetic 
and general reasoning tasks (GSM8K 65\%, AQuA 48\%, SVAMP 77\%).

\begin{figure}
  \begin{center}
    \includegraphics[width=10cm]{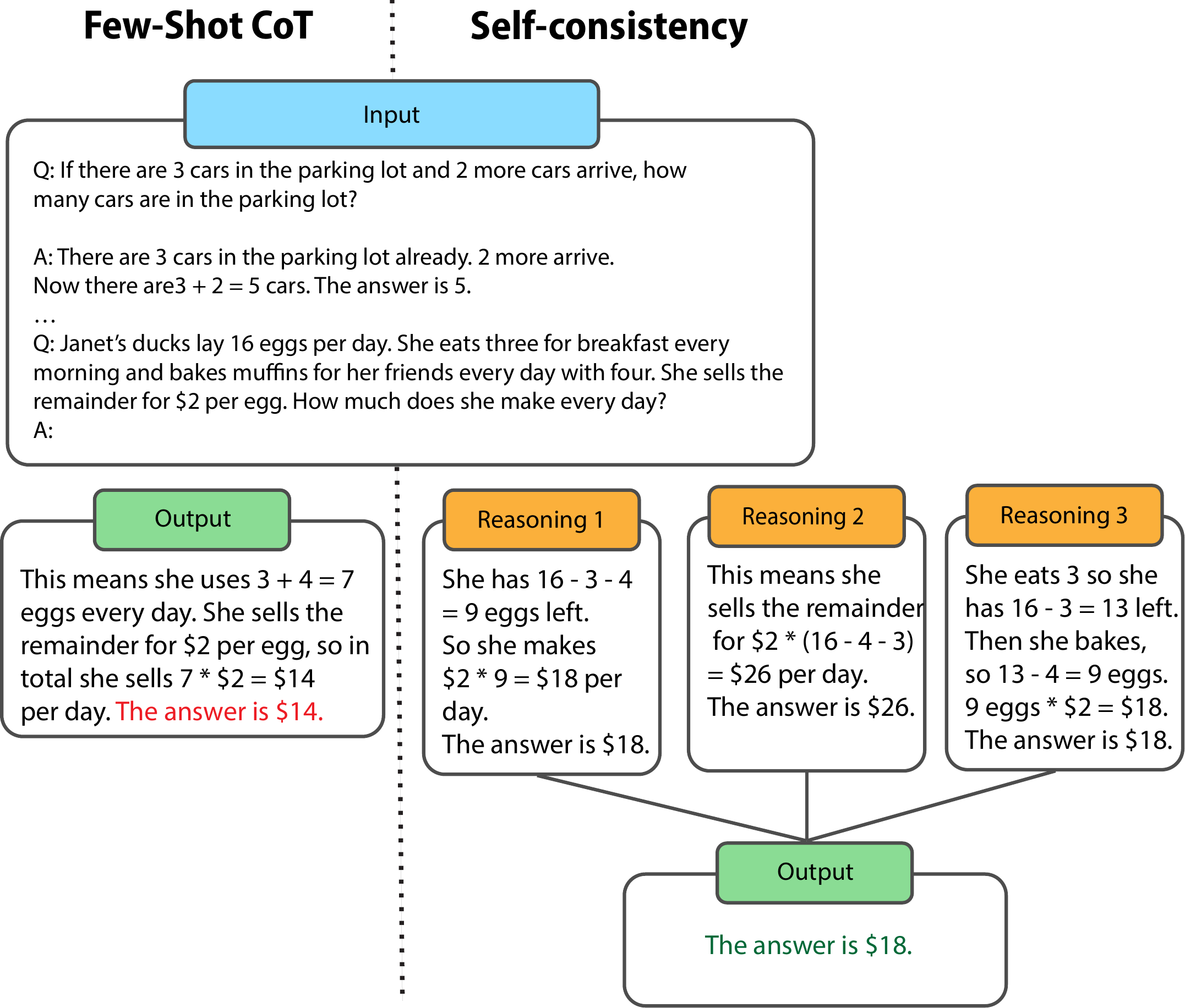}
    \caption{Self-Consistency (Adapted from \citep{wang2022self}),  showing majority voting over answers that the model produces}\label{fig:sc}
  \end{center}
\end{figure}

A popular related approach is  Self-consistency \citep{wang2022self}.\label{selfconsistency}
Self-consistency is a straightforward ensemble approach (a well-known machine learning technique to make a strong learner out of multiple weaker
learners \citep{sagi2018ensemble,breiman2001random}).
Greedy single-path decoding is replaced 
by sampling diverse reasoning paths, evaluating them, and selecting the most consistent answer. Self-consistency asks the LLM to simply perform the same query multiple times, and takes the majority-vote of the answers, or decoding paths. Self-consistency
works since complex reasoning problems typically
allow different reasoning paths that lead to the 
correct answer. Figure~\ref{fig:sc} summarizes the approach.
Self-consistency has been evaluated on arithmetic reasoning, commonsense
reasoning and symbolic reasoning, on a variety of LLMs, including GPT-3
\citep{tay2022ul2,brown2020language,thoppilan2022lamda,chowdhery2023palm}.
Self-consistency further improves the performance of Chain-of-thought
 by 10-20 percentage points, and has been used as a baseline in many of the other approaches in this survey.

\subsubsection{Tool-based Validation}

Another approach to improve the accuracy of evaluating the reasoning steps is to switch from a natural to a formal language. The advantage
of a formal language is that it is less ambiguous than a natural language. Examples are computer languages, such as Python, or mathematical equations. Using a formal language for reasoning is a popular approach, and we discuss seven
papers.\footnote{In addition to the languages discussed in this survey, there is a long tradition in AI of using formal reasoning based on logic. The interest in Chain-of-thought has also stimulated  research into external symbolic logic solvers that are called by the LLM  \citep{pan2023logic,xu2024faithful, saparov2022language,tafjord2020proofwriter, han2022folio,srivastava2022beyond,zhong2022analytical}, or where LLMs transform the problem in a logic problem. The use of LLMs for generating formal languages is further discussed in section~\ref{sec:proglang}.} 
Many approaches  generate the steps in 
Python, and the code can then be evaluated by a formal
evaluator, such as a compiler, debugger, or interpreter.

LLMs have been quite successful in generating computer code from natural language prompts.
\citet{chen2021evaluating} introduced Codex, a GPT model that
was trained on publicly available  code in the repository
GitHub. A production version of this work was introduced under the
name GitHub Copilot. Codex is often able to generate correct programs from
 descriptions in natural language, such as commentary strings.

The work on Codex is used as a basis for further research on reasoning in LLMs.
Human programmers, when writing code, typically
follow a cycle of writing some code, executing it to look for errors,
and then using the feedback to improve the code. This step-by-step approach
is followed in Self-debugging 
\citep{chen2023teaching}. 
It follows the same steps of  code generation, 
code execution, and  code explanation.
%
 %
Self-debugging is able 
to identify mistakes in its own
code 
by investigating the execution
results, and can also provide an explanation of the generated code, in
natural language. It achieves strong performance: the text-to-SQL Spider benchmark improves by 9 points,  and the C++ to Python Transcoder benchmark improves by 12 percentage points.



Several works  generate working code tuned for solving specific problems automatically, without human feedback.
\citet{romera2024mathematical} introduced FunSearch, an approach that integrates formal methods and LLMs to enhance mathematical reasoning and code generation. FunSearch 
uses a genetic  approach with multiple populations of candidate solutions (programs), which are  evaluated using a function depending on the problem specification. In addition to the 
evaluation function, also an initial program is given to the LLM in the first prompt. After evaluating a number of generated programs from the starting prompt, a new prompt  is created, in an iterative fashion, combining a selection of  sampled programs sorted  according to their evaluation score, and the LLM is requested to generate a new program.
Another work  leverages evolutionary computation methods to generate and optimize evolutionary algorithms \citep{van2024llamea}. This approach, LLaMEA (Large Language Model Evolutionary Algorithm),  utilizes LLMs to  design and optimize  evolutionary algorithms. The approach uses LLMs to generate initial algorithmic structures, which are then refined through  mutation and selection. This  enhances the efficiency of algorithm design, particularly in fields requiring innovative and adaptive solutions, improving accuracy on the {\em Black-Box Optimization Benchmark} suite \citep{hansen2010black} (BBOB)   by 11 percentage points.
A key difference between FunSearch and LLaMEA is that LLaMEA uses a sample-efficient elitism strategy by iteratively improving the best-so-far solution, requiring significantly fewer prompt evaluations than the large-population strategy proposed in FunSearch. 
Evolutionary approaches let the LLM  discover new algorithms, solving existing problems in new ways, or solving entirely new problems. Another method, Evolution-of-heuristics  \citep{liu2024evolution}, was proposed for evolving code snippets for guided local search to solve combinatorial optimization problems, such as the Traveling Salesperson Problem.

To improve prompt-based reasoning, Codex is used in 
an  ensemble approach named  MathPrompter
\citep{imani2023mathprompter}. This approach   generates multiple algebraic
expressions or Python functions, which then 
solve the same math problem. The results are
compared, just like in Self-consistency and
Self-verification, raising the
confidence level in the results.
MathPrompter achieved 
state-of-the-art accuracy on the MultiArith dataset (78.7\% $\rightarrow$ 92.5\%), evaluated on GPT-3 175B.

Two other approaches that use a formal language are Program-of-thought  \citep{chen2022program} and Program-aided-language  \citep{gao2023pal}. Both approaches use the LLM to generate Python and then use an interpreter to evaluate the result.
The approaches are similar although Program-aided-language uses generic prompts, and has been tested on more benchmarks. 
Figure~\ref{fig:pal} (fourth column) illustrates the Program-aided-language approach.
%
%
When the evaluation of the reasoning steps is off-loaded  to the Python
interpreter, decomposing the natural language problem into executable
code steps 
remains the only task for the LLM. (Earlier work in math word problems showed how to decompose a problem and reach
an answer \citep{ling2017program}.) 
\citet{gao2023pal} provide extensive experimental
evidence about the synergy between the neural LLM and the symbolic interpreter.
Experiments are performed on 13 mathematical, symbolic, and algorithmic reasoning
tasks, achieving more accurate results than much larger
models (for example, Program-aided-language reported 72\% on GSM8K and  93\% on Penguins). 

\subsubsection{External Model Validation}
We have seen many  examples  of successful prompt-based
in-context reasoning and evaluation (at inference time---where no parameters were changed). We will now look at  reasoning
approaches that follow a more traditional parameter training approach. 
All approaches evaluate the output of the model and generate
corrective data. That data is then added to the training pipeline, and the model is subsequently finetuned.

\paragraph{Finetuning}
The Refiner approach \citep{paul2023refiner} uses a
generator model and a critic model  that provide fine-grained feedback on reasoning
errors.
The generator generates multiple reasoning hypotheses, and the critic evaluates results by randomly selecting a hypothesis for feedback. The generator model is then finetuned based on its reasoning errors. A small supervised model is used to overcome the cold-start problem.
%
The  approach  achieves 78\% accuracy on GSM8K and also works well on related problems.

\citet{welleck2022generating}  follow a similar approach, which they call
Self-correction. Here, the corrector is a separate model specialized in refining the outputs of the generator. Unlike Refiner, where the generator is finetuned based on the critic, Self-correction  finetunes the corrector to rectify errors in the hypotheses produced by the generator. Self-corrector is not applied to math word problems, but to program synthesis, where a small corrector reduces toxicity to 0\%.

%

A third finetuning approach is Self-improvement \citep{huang2022large}. Here,
too, the base model data is augmented by LLM-generated rationales, and then finetuned. The authors achieve 82\% accuracy on GSM8K and similarly high scores on question answering and adversarial benchmarks.
Noteworthy in all three finetuning approaches is that LLMs are
capable of improving themselves by training on their own generated
output, and that stability problems from feedback loops are overcome.

\paragraph{Dataset Augmentation}
The final finetuning approach that we discuss uses dataset
augmentation. In Self-taught-reasoner \citep{zelikman2022star}, an
intermediate reasoning is generated, called a {\em rationale}. 
Rationales are shown to be valuable 
across diverse tasks such as
mathematical and commonsense reasoning, code evaluation, social bias
inference, and natural language inference. First an augmentation dataset is created
by attempting to solve the original dataset.
Next, the dataset is augmented using rationalizations and ground-truth answers to problems the model failed to solve. Finally, the model is finetuned on the combined dataset.
%
Self-taught-reasoner performs comparably (72\%) to finetuning a 30 times larger model on CommonsenseQA.
%
%
%


\paragraph{Reasoning about Robot Behavior}
In addition to math word problems, computer code, and common sense, prompt-based reasoning has also
been used to improve robot behavior.
Language models contain a large amount of information about the
real world \citep{ahn2022can}. In theory, this should allow them to reason 
realistically  about robotic behavior. However, the models do not
have knowledge about specific embodied aspects of a particular
robot. If we could compare a Scratchpad-like list of intermediate reasoning
steps with a list of possible movements of the robot in its
environment, then we could prevent the model from suggesting impossible joint movements and actions, and prevent failures. 

Say-can \citep{ahn2022can} learns a value function \citep{kaelbling1996reinforcement} of the
behavior of a robot and its environment using 
temporal difference reinforcement learning \cite{sutton1988learning}. This value function is 
combined with prompt-based reasoning by the language model, to
constrain it from suggesting impossible  actions. 
%
%
The goal of Say-can is to ground language in robotic
affordances. In contrast to Scratchpad, which used supervised
learning, the affordance model is learned interactively by reinforcement
learning, and then applied using prompt-based learning on
the LLM. 
The language model has high-level semantic knowledge about the
task 
({\em Say}). 
The learned affordance function ({\em Can})
provides an environment-grounding on what is possible. 
Say-can achieves a 31\% success rate  on 101
real-world robotic kitchen tasks.


Where Say-can learns affordances as a separate function, Inner-monologue \citep{huang2022inner} formulates robotic
planning directly as part of the language prompt, internally.
%
%
The input consists of many elements:  textual descriptions from InstructGPT
\citep{brown2020language} for multi-step planning, scripted modules
for object recognition,  success detection,  task-progress scene
description, and language-conditioned pick-and-place primitives,
similar to CLIPort \citep{shridhar2022cliport}. 
%
%
%
The language feedback that is thus generated significantly improves
performance on three benchmarks, achieving 90\% accuracy on simulated and real table top
rearrangement tasks and  60\% on the kitchen environment. 
There are many other studies into robotic behavior. An approach related to Inner-monologue is Chain-of-tools, which proposes a plan-execute-observe pipeline to ground reasoning about tool behavior \citep{shi2024chain,shi2024learning}.

This concludes our discussion of the second stage of the reasoning pipeline, {\em evaluation} of the
reasoning steps.

\subsection{Control of Steps}
The third stage is  {\em control}. This stage
controls how many sub-steps are generated, and how deep into the
future the reasoning chain is generated.
There are three main approaches: (3.3.1) {\em greedy selection}, which
generates a step and then follows it, (3.3.2) {\em ensemble strategy}, which generates
a set of possible next steps, and (3.3.3) a  ({\em reinforcement learning}) search which
generates multiple options for the steps, 
traversing a search tree with backtracking,
controlling an exponential search space \citep{xu2025towards}.  

\subsubsection{Greedy Selection}
Most earlier works on prompt-based reasoning follow the greedy approach:
generate a single prompt with a single sequence of steps and follow them.
Among the greedy reasoners are Chain-of-thought, Auto-CoT, and
Zero-shot CoT. Inner
Monologue and Say-Can also use greedy reasoning. 

In Least-to-most prompting
\citep{zhou2022least}, the key idea is to break down a
complex problem into simpler subproblems and then solve
these in sequence, 
explicitly encoding them in the prompt, related to Complexity-based prompting. In Least-to-most
the 
answers to previously solved subproblems help in finding the answer, as  a curriculum \citep{bengio2009curriculum}.
On symbolic manipulation, compositional
generalization, and math reasoning,  Least-to-most
prompting generalizes  well, achieving 99\% accuracy on a compositional generalization benchmark. 


\subsubsection{Ensemble Strategy}
The second kind of reasoning control is based on an
ensemble of (sequences of) reasoning steps.  For most problems, multiple different
options exist for the next step. When all or some of these are
generated and evaluated, then the consensus result
can be reported as the outcome of an ensemble of steps. 
Self-consistency \citep{wang2022self} and Self-verification \citep{weng2022large} (in
Section~\ref{selfconsistency}) are popular ensemble approaches to
evaluate the results of reasoning steps, 
in which   greedy single-path decoding used in
Chain-of-thought prompting is replaced 
by  a diverse set of  paths.
%
%
%
Taking this further, Chain-of-experts uses a mixture-of-experts
ensemble for complex combinatorial  problems \citep{xiao2023chain}.
Program-aided-language and  MathPrompter also use the ensemble approach. 
The ensemble approach is popular  in reasoning with LLM.

\subsubsection{Reinforcement Learning}\label{sec:self-reflection}
In reasoning, often multiple valid  steps are possible,
but pursuing all possibilities over multiple trajectories
may lead to an infeasible number of possibilities.
The third kind of reasoning control is to use a full-fledged
controller that can traverse a tree,  or  perform reinforcement
learning to do so \citep{sutton2018reinforcement,kaelbling1996reinforcement,plaat2022deep}. 
When decomposing the problem, multiple alternative steps are
generated that can be searched multiple steps into the future. Then, backtracking can be performed, allowing alternative steps to be tried.

Where greedy and ensemble processes can be controlled with a prompt by the LLM, this third category is more complex, and an external algorithm is used to control the reasoning process. The external algorithms call the LLM as a subroutine prompting it to perform requested tasks. The external algorithm allows more complex reasoning control, but we are now beyond prompt-based self-reasoning: control has been given to an  algorithm that is external to the LLM and external to prompt-learning.

\begin{figure}
  \begin{center}
    \includegraphics[width=14cm]{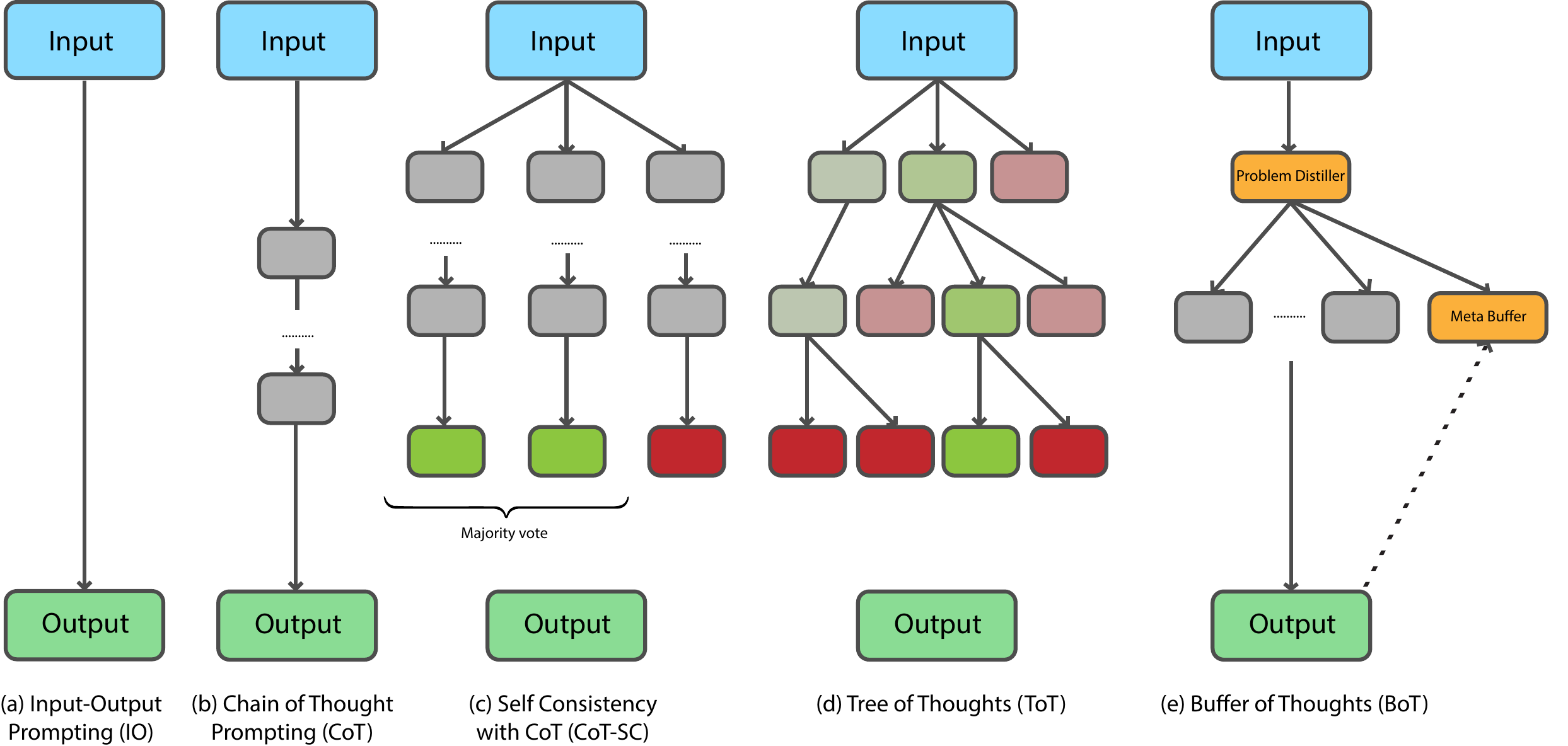}
    \caption{Reasoning structure of Simple prompting (a), Chain-of-thought (b), Self-Consistency (c), Tree-of-Thoughts (d) and Buffer of Thoughts (e); Prompts in (a) consist of an instruction, and, perhaps, few-shot examples of answers, providing no reasoning guidance; Chain-of-thought (b) guides the model with few-shot examples of reasoning steps; Self-Consistency (c) takes the majority vote of existing answers in the model; Tree-of-thoughts (d) uses an  external prompt-optimizing algorithm to guide  the model to perform a tree search over reasoning steps, exploring multiple alternatives; Buffer-of-thoughts (e) uses a problem distiller that  stores high-level thoughts distilled from different problems in a meta-buffer, which  is actively managed for capacity} \label{fig:tot} \label{fig:bot}  
  \end{center}
\end{figure}

We start our discussion of control strategies with depth-first and breadth-first search, then go to beam search, and then to full reinforcement learning.
%
\label{sec:buffer}
A complex reasoning space can be traversed with a search algorithm.
Tree-of-thoughts \citep{yao2024tree} uses breadth-first or depth-first search  to dynamically follow different reasoning steps. 
%
Tree-of-thoughts calls the LLM with two different types of prompt: to generate the sub-problems, and  to evaluate them. 
Together, the trio of generation (LLM-prompt-1), evaluation (LLM-prompt-2), and control (search-algorithm)
allow systematic exploration of the space of reasoning steps
with look-ahead and backtracking. 
The authors compare their approach to Chain-of-thought and Self-consistency on the Game of 24, Creative writing, and Mini crossword, achieving an accuracy of 74\% on a Game of 24 task. The other tasks are evaluated qualitatively. 
Figure~\ref{fig:tot} illustrates the different reasoning structures.\footnote{A similarly named approach is Graph-of-thoughts \citep{besta2024graph}. Graph-of-thoughts allows more general reasoning graphs, providing a formal framework, where the different elements can then be specified manually.}

Another approach, Buffer-of-thoughts \citep{yang2024buffer}, goes a step  towards metareasoning \citep{gao2024meta}. It introduces a 
meta-buffer that stores  high-level {\em thought-templates}. These  thought-templates are derived from a variety of tasks by a problem distiller. They  are then instantiated for specific tasks. To address size restrictions of the LLM context window, the meta-buffer is managed for capacity and thoughts can be combined. Figure~\ref{fig:bot} compares the Buffer-of-thoughts approach to other approaches such as Chain-of-thought and Tree-of-thoughts. Buffer-of-thoughts outperforms other methods in puzzles such as Game of 24 (by 11\%) and checkmating (by 51\%). 
Thought templates are related to metacognition (thinking about thinking), which is further discussed in Section~\ref{sec:metacog}. 


A related search method is 
Beam-search-for-reasoning \citep{xie2024self}. 
When the space of possible reasoning paths is  large, 
Beam-search 
searches a promising part of this space. It uses self-evaluation  to control exploration and to evaluate (decode) reasoning steps. 
%
Beam search
uses Program-aided-language models for math word problems \citep{gao2023pal}.
Using a Codex backbone \citep{chen2021evaluating}, it surpasses the few-shot baselines by 6.34\%, 9.56\%, and 5.46\% on the GSM8K, AQuA, and StrategyQA benchmarks, respectively.

Reinforcement learning is another step in the sophistication of optimization algorithms. 
It learns by interactive sampling, improving
its policy based on rewards from the environment  \citep{sutton2018reinforcement}.
To use reinforcement learning, the reasoning problem is formulated as a Markov Decision Process: the agent-algorithm creates a prompt for a next step (an {\em action}), to sample a step ($t$) and get an answer ({\em state}, {\em reward}) from the environment-model. 
The answer can then be used to improve the prompt (next action), 
using the rewards to improve its policy of best actions for each state. The approaches that use reinforcement learning also do so in the form of an external algorithm. 

Progressive-hint-prompting (PHP) uses  reinforcement learning to interactively improve prompts \citep{zheng2023progressive}. 
%
%
PHP 
calls the LLM with dynamic prompts, using previously generated
answers as hints, to progressively prompt the LLM towards the correct
answers. It works as follows: (1) given a question (prompt), the LLM
provides a base answer, and (2) by combining the question and answer, the LLM is 
queried and obtains a subsequent answer. We (3) repeat
operation (2) until the answer becomes stable, as a regular policy-optimizing reinforcement learning algorithm.
The authors have combined PHP with Chain-of-thought and with
Self-consistency.
Using GPT-4, state-of-the-art performance was achieved on grade 
school
math questions (95\%), simple math word problems (91\%) and algebraic
question answering (79\%).

Another approach that is motivated by improving answers from feedback, is 
Self-refine \citep{madaan2023self}. 
Like PHP, the LLM generates an initial output and provides feedback for
its answer, using the LLM to refine itself, iteratively. 
%
Self-refine prompts the LLM in three ways: (1) for initial generation, (2) for
feedback, and (3) for refinement, following a greedy reasoning chain. Self-refine has been used with
GPT-3.5 and GPT-4 as base LLMs, and has been benchmarked beyond math word problems on dialogue
response generation \citep{askari2024self}, code optimization, code readability improvement,
math reasoning, sentiment reversal,  acronym generation, and
constrained generation, showing substantial improvements over the base
models (typically around 30 percentage points, see Table~\ref{tab:geneval2}).

Another approach that combines reinforcement learning and LLMs is ReAct \citep{yao2022react}. However, ReAct does so in a different way.
Most works  focus on reasoning by the LLM, not on actions by an agent.
%
The goal of  ReAct 
is to combine progress in reasoning with action plan generation. (Or, to put it differently, other approaches  use RL to improve LLM-reasoning, ReAct uses LLMs to improve RL agent policies.)
ReAct uses Chain-of-thought prompt-learning as part of an RL framework that also uses external knowledge sources (Wikipedia) and finetuning; for error reduction, grounding, and for reducing hallucination. The framework allows hand-written prompts.
%
%
On two interactive decision
making benchmarks (Alfworld and WebShop),
ReAct outperforms imitation
and reinforcement learning methods by an absolute success rate of 34\%
and 10\% respectively.

\begin{figure}
  \begin{center}
    \includegraphics[width=8cm]{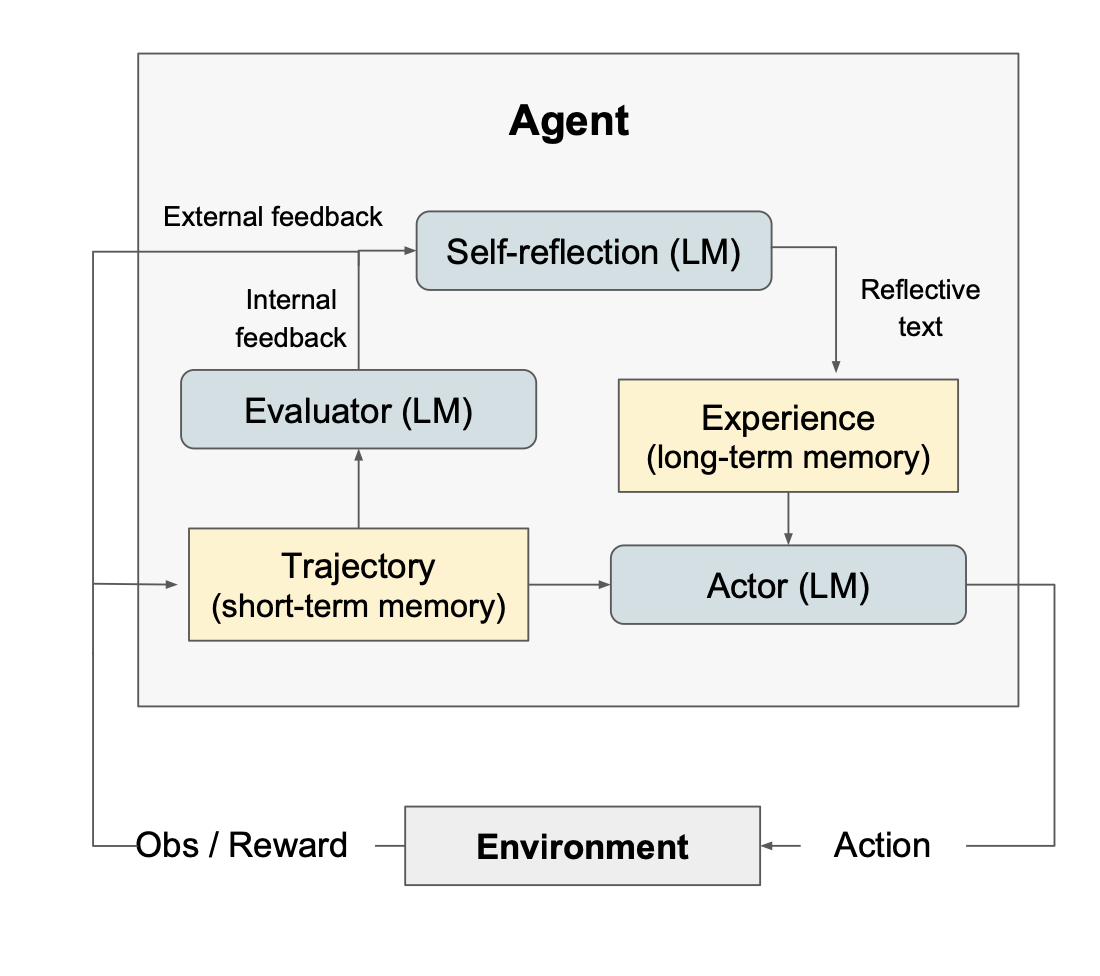}
    \caption{Architecture of Reflexion  \citep{shinn2024reflexion}, showing a close resemblance to the agent/environment structure of reinforcement learning \citep{sutton2018reinforcement}}\label{fig:reflexion2}
  \end{center}
\end{figure}

The ReAct work has been developed further.
Reflexion \citep{shinn2024reflexion} 
creates AI agents that learn by reflecting on failure and
enhance their results, much like humans do. 
Reflexion 
uses three language
models: actor, evaluator, and reflector.
It works as follows: (1) an actor  generates text and actions,
(2) an evaluator model  scores the outputs produced by the actor,
and (3) a self-reflection model  generates verbal reinforcement
cues to assist the actor to self-improve (see Figure~\ref{fig:reflexion2}).
For the actor, Chain-of-thought
and ReAct  can be used. Reflexion is evaluated on
decision-making, reasoning, and coding tasks. Improvements of 10-20
percentage points are reported.

The reinforcement learning approaches that we discussed so far---React, Self-refine, Tree-of-thoughts, Buffer-of-thoughts, Reflexion---use external algorithms to manage state for the prompt improvement loop. A more elegant solution would be to perform reinforcement learning fully in-context, within-prompt. Indeed, \citet{krishnamurthy2024can} explicitly asked the question whether LLMs can explore in-context.  This is the goal of the Algorithm-of-thoughts (AoT) approach \citep{sel2023algorithm}. Following work by \citet{lee2023supervised,wang2024transformers,demircan2024sparse} that showed that transformer architectures can be pretrained and finetuned to perform in-context reinforcement learning, AoT aims to do so in general LLMs such as GPT-4, Claude and Gemini 1.5.   They achieve results comparable to Tree-of thoughts on GSM8K, StrategyQA, and Crosswords. Achieving these results with in-context RL  is a promising result for in-context learning. Other search-like in-context algorithms are studied by \citet{schultz2024mastering} and \citet{kempinski2025game}. Further works in games find that LLMs struggle with the difference between generating an algorithm for a problem, and executing that algorithm correctly (the {\em knowing-doing gap} \citep{paglieri2024balrog,ruoss2024lmact}), and struggle with the difference between using and mentioning game concepts \citep{van2025baba,cloos2024baba}. Work on implicit reasoning is ongoing, \citet{li2025implicit} provide a survey.

\begin{figure}  \begin{center} \includegraphics[width=10cm]{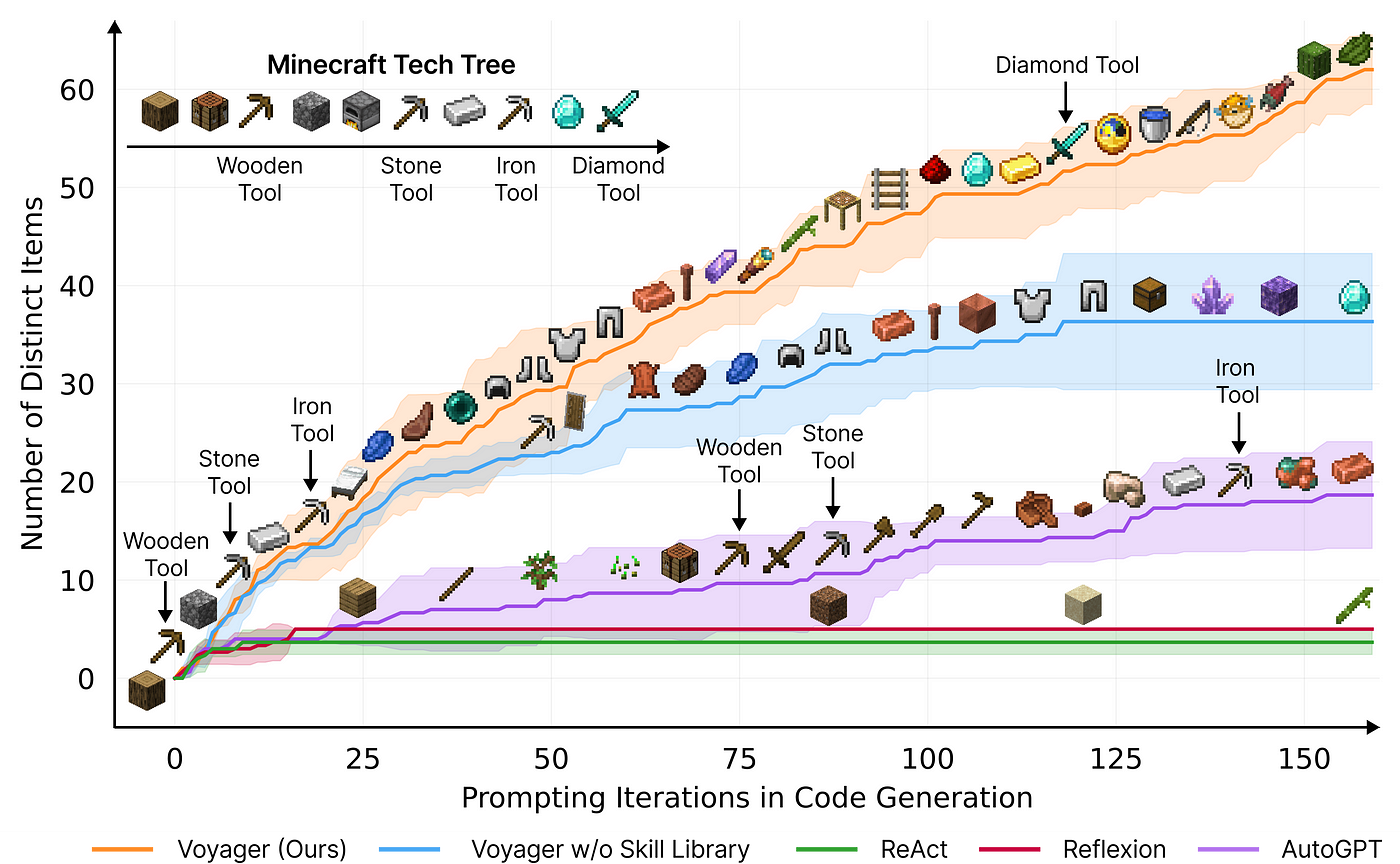} \caption{Performance of Voyager in Minecraft  \citep{wang2023voyager}; Voyager performs well, 
reaching high scores  by acquiring many tools}\label{fig:minecraft} \end{center} 
\end{figure}

To conclude this overview of reinforcement learning approaches, we discuss an application in the games domain.
Voyager \citep{wang2023voyager} is an agent for the game of Minecraft that uses an iterative
prompting mechanism that generates code for embodied control. The
agent includes self-verifcation  and a skill library 
to
maximize exploration. 
The goal  is to discover  diverse items in Minecraft, a form of novelty search
\citep{eysenbach2018diversity}. Voyager performs well, 
reaching high scores  by acquiring many tools (see Figure~\ref{fig:minecraft}) 15 times faster than the baseline. 

The applications of reinforcement learning in LLM reasoning are many, and the connections run deep \citep{pternea2024rl}. \citet{wang2024offline} use the similarity between RL timesteps and LLM reasoning steps to jointly train a value function together with the
LLM policy by optimizing the  Bellman equation, achieving 85\% accuracy on GSM8K and 81\% on Alfworld. \citet{guo2025deepseek,team2025kimi} replace supervised finetuning by reinforcement learning, integrating it with  reasoning, achieving large efficiency gains in the training pipeline in the DeepSeek r1 and Kimi models.


\section{Discussion}\label{sec:discussion}
We have reviewed many reasoning approaches. 
It is now time to reflect,
to discuss limitations, and to look for promising areas of future work. 
First we  discuss how to choose an approach for an application. Next, we discuss issues concerning hallucination, faithful reasoning, and scaling. Then we discuss what LLMs can and cannot do. We highlight connections with sequential decision processes, metacognition, and programming languages, and we end with a research agenda.

\subsection{Matching Methods and Applications}\label{sec:guidance}
This survey has discussed  different approaches, applications, and benchmarks. It is often difficult to directly compare benchmark results between individual papers, due to differences in measurement setup. For example, \citet{kamoi2024can} critically compare self-reflection studies. Despite the challenges in comparing benchmark results, we can provide general guidance on what approach to use for different types of applications.

For {\em simple reasoning situations}, that require a single sequence of reasoning steps, Chain-of-thought is typically  used. When the error rate becomes too high, verification methods such as Self-consistency are popular. Indeed, this combination is used in many modern reasoning LLMs  such as DeepSeek and GPT-o1 \citep{guo2025deepseek,team2025kimi,lambert2024tulu,huang2024o1}.

For {\em combinatorial puzzles and games} that would traditionally require backtracking, tree-search or reinforcement learning solution methods must keep track of the enumeration state. However, an LLM without an external algorithm often fails to keep track of this state correctly between calls to the model. An approach such as Tree-of-thoughts,   or ReAct, should be considered. Here,  the state is managed externally, and the LLM is called for application-dependent functional processing.  Various in-context reinforcement learning methods are being developed where state is managed in-context, as in, for example, Algorithm-of-thoughts. When performance is not sufficient,  finetuning or pretraining for the domain at hand can  be used.

For {\em complex problems}, where a sequence of steps, or external control, fail to perform well, other reasoning approaches are needed. 
The LLM can be asked to generate a program for the problem. In this  approach,  the LLM acts as a coding agent for its user. The LLM generates a program or a sequence of commands, which are then executed by an external tool, such as an interpreter, a robot, a planner, or a logic solver. In robotics, vision-language-action models have achieved impressive results \citep{kim2024openvla}, but also strong results in logistics \citep{bohnet2024exploring}, finance and medicine are reported \citep{plaat2025agentic}.

\subsection{Hallucination, Faithfulness and Scaling}
Reasoning by LLMs is not error-free, and many studies 
aim to provide deeper insight into  the reasoning processes in  language models.
\citet{saparov2022language} introduce a synthetic question/answer
dataset designed to evaluate the reasoning abilities of LLMs. The work
showed that LLMs are capable of reasoning to a certain degree, but that
Chain-of-thought struggles with proof trees with a wide branching factor.
In another study, \citet{wang2022towards}  aim to increase our
understanding of how Chain-of-thought works. The authors find that 
the order of the reasoning steps
is important. Prompts should be relevant to the
question, and coherent (steps should be in the correct order).
\citet{jin2024impact} also study the impact of reasoning step length on
LLMs, again finding a strong positive correlation between the length of the prompt and reasoning abilities.
%
%
Next, we discuss  works on errors  in the Chain-of-thought approach, studying whether the reasoning of the LLM is faithful, or that it gives the right answer for the wrong reason.

\subsubsection{Faithfulness}
Chain-of-thought approaches prompt a language model to take
certain steps to solve the problem that the prompt specifies. One
can ask the question whether  those steps are indeed the steps
that the model has followed (faithful reasoning) or whether it took another road to arrive
at the same answer (unfaithful reasoning).
A few studies measure the faithfulness of reasoning with LLMs. 
\citet{lanham2023measuring} notes that just like organic reasoners,
a model's reasoning may be post-hoc, it may be constructed after a
certain conclusion has been found. By deliberately adding mistakes to
the chain of thought, the authors measure the faithfulness of the
model. They find a wide variation of post-hoc reasoning, with a
tendency of larger models to be less faithful. Like regular LLMs, when not properly grounded, (Chain-of-thought) reasoning  suffers from hallucination \citep{huang2023survey}.

Another study adds deliberate bias to the prompt. For example, in a multiple-choice setting, they always
make answer (A) the correct answer
\citep{turpin2024language}. They find that a bias towards wrong
answers can cause significant drops in accuracy, and that models frequently
generate Chain-of-though explanations rationalizing  wrong
answers.
%
%
To address issues of faithfulness, 
\citet{lyu2023faithful,xu2024faithful} propose Faithful-chain-of-thought. This
approach involves two stages. First, the natural language query is
translated into a formal symbolic language. Second, the problem-solving
stage processes the formal language, and can explain the reasoning
steps it has thus taken. For the symbolic language, Python,
Datalog, or PDDL is suggested. Another approach, mechanistic
interpretability, studies methods to target individual representations inside the LLM, to see if the expected behavior occurs in practice \citep{rai2024practical,bereska2024mechanistic,chen2025does}.

Faithfulness studies tell us more about how models reason. Further surveys on this topic are \citet{mondorf2024beyond,chuang2024large,luo2023reasoning,paul2024making},


\subsubsection{Scaling} 
The emergent abilities of LLMs have prompted research into the nature of scaling and reasoning with LLMs, and, specifically, how reasoning capabilities can be transferred to smaller language models. Scaling laws of LLMs are an active area of study, see for example \cite{kaplan2020scaling,henighan2020scaling,hoffmann2022training}.
Distillation of reasoning to smaller models can work surprisingly well in situations with more explicit instructions, and given the computational cost of training LLMs, there is much
interest in transferring knowledge to small language models.  Comprehensive surveys on knowledge distillation are
\citet{xu2024survey,gu2023minillm}. 
For reasoning specifically, \citet{magister2022teaching} have studied
reasoning in small language models, using a student model that learns
from a teacher model, by finetuning. 
Other works focus on prompt distillation for retrieval \cite{dai2022promptagator}, recommendation \citep{li2023prompt},  embodied agents \citep{choiembodied}, and LLM graph reasoning \citep{zhang2024can}.  
%

Another study related to Self-taught-reasoner
focuses on
explanation in small language models, also achieving good results of knowledge transfer \citep{li2022explanations}.

\subsection{Limitations: What LLMs Can and Cannot do}
The capabilities of LLMs are impressive.  LLMs can  be seen as large text-based surrogate models of the world (or the world how we describe it on the internet), and thus allow   reasoning 
about a large variety of contexts and problems. Reasoning,  such as in math word problems, were one of the capabilities that LLMs could not achieve, until recently. Let us look more closely at what language models currently can and cannot do.

\subsubsection{What Can LLMs Do?}


With the right prompt, LLMs are able to solve many of the  problems in grade 
school math word benchmarks and beyond. 
Prompt-based learning is able to perform  reasoning tasks such as math word problems, robotic movement gaming, and Python code generation, at inference time, without expensive parameter training. 

A taxonomy of generate-evaluate-control is able to describe the structure of the current LLM reasoning literature.
Furthermore, the accuracy of the reasoning chains can be improved with ensemble methods, and self-verification.
Hallucination can be reduced by grounding the model with external models, such as for robotic affordances, and information retrieval from search engines and Wikipedia. 
Going a step further, using external  control algorithms   as scaffolding (such as search or reinforcement learning), dynamic prompts can use the LLMs to perform  complex and dynamic reasoning patterns.
Note that the reasoning control is now outside the core LLM: an external control algorithm, on top of in-context-learning,  dynamically generating  prompts for the LLM. 

At this point, it is interesting to note the confluence of  two schools of thought in artificial intelligence: symbolic and connectionist.\footnote{Reasoning and planning have been studied since the start of 
artificial intelligence, starting with logic and reasoning \citep{newell1961computer},
search algorithms in puzzles and board games \citep{korf1999artificial,plaat2020learning,plaat1996research},
robot planning \citep{fikes1971strips}, classical machine learning such as
decision trees and support vector machines
\citep{flach2012machine,breiman2001random,cortes1995support}, through knowledge
representation and the semantic web \citep{van2008handbook}. Ever since the success
of the connectionist approach \citep{brooks1990elephants,lecun2015deep,goodfellow2016deep} (deep learning, including
LLMs) researchers have tried to join the two approaches \citep{tesauro1995temporal,silver2017mastering,wang2023analysis}.
}
Search and reinforcement learning are rooted in the symbolic tradition, while LLMs and deep learning are rooted in the connectionist tradition. The literature in this survey combines the two traditions. Higher performance reasoning is created with a (symbolic) searcher/learner on top of a (connectionist) LLM. In other fields similar combinations can be seen (for example, AlphaFold \citep{bryant2022improved,jumper2021highly} and retrosynthesis of molecules \citep{segler2018planning}). The LLM helps ground symbolic reasoning methods in language; symbolic  methods help create prompts that let the LLM perform dynamic reasoning. 

We   note that benchmarks such as GSM8K have been central for the progress of the field, and that while reasoning started with math word problems, the field has extended to robotics, autonomous agents, games, and most emphatically computer code. Formal languages  play an important role in the intermediate multi-step reasoning chains. 


\subsubsection{What Can LLMs Not Do?}
Now that grade school math word problems are largely solvable, harder reasoning benchmarks in other domains are appearing \citep{ahn2024large}. 
Most of the reasoning capabilities exhibited by LLMs are due to the  representational powers of the transformer architecture and how in-context learning is able to harness them. Prompt engineering and prompt control play a crucial role in the   reasoning that we have seen in this survey.
Models can be instructed to write their own reasoning prompts; however, such Auto-GPT or Auto-CoT prompts are prone to feedback-loop problems, and need careful evaluation, verification, and grounding in the real world, to prevent degeneration into a hallucinatory world of their own.
Models can also be instructed to interact with the world, and become the tool of external scaffolding   that evaluates, controls and improves the prompts \citep{plaat2025agentic}. 
Some of what we experience as reasoning {\em by} the LLM, is controlled by the prompt or the scaffolding algorithm. Studies into in-context reinforcement learning aim to answer the question if prompt learning is able get the LLM to create a prompt to exhibit dynamic reasoning by itself \citep{lee2023supervised,schultz2024mastering,demircan2024sparse,paglieri2024balrog}.

Some studies on symbolic planning are  critical on the abilities of LLMs \citep{valmeekam2023planning}, and show examples of planning failures,
arguing that LLMs
are better used  to improve heuristic elements of traditional planners, such as PDDL \citep{kambhampati2024llms},
to strengthen traditional symbolic planning approaches. 

Other works study the dangers of the size of LLMs. \citet{bender2021dangers} mention the environmental risks associated with the large computational training demands, as well as the difficulty of understanding the training data, for example in the context of bias.  Furthermore, there are ethical, legal, and copyright concerns regarding the data that LLMs are trained on. Finally, to prevent putting too much trust in the outcome of LLMs, we should understand their failure modes better, such as the well-publicized problems of hallucination (inventing facts that look right but are not).
Here, mechanistic interpretability can be used to explore LLM representations and understand where they go wrong \cite{bereska2024mechanistic,rai2024practical,chen2025does}.

Some of the names of the approaches surveyed in this paper are  suggestive of self-awareness and self-reflective capabilities. 
True (human) self-reflection, or metacognition, is still largely outside the capabilities of current LLMs. LLMs can be prompted to reason, to take small steps, to self-evaluate, and their search process can be controlled by an external algorithm. The self-reflective type of ``intelligence'' is written into the prompt by the prompt engineer or the control algorithm. We are unaware of any LLM that has been made to reflect on, or even control, its reasoning processes, controlling how many reasoning steps it should take, or limiting its reasoning once the answer had become good enough. 
%
%
True  self-reflection remains future work, although some steps have been taken, as we will  discuss next.

\subsubsection{Reasoning towards Metacognition}\label{sec:metacog}

Human thought  exhibits the ability to reason about self,
we are able to think about our own thinking processes. Metacognition
studies this phenomenon \citep{veenman2006metacognition}. Prompted by the success of Chain-of-thought and related works,
metacognition has also been studied in the context of LLMs~\citep{toy2024metacognition}.

Many reasoning approaches highlight self-reflective aspects 
in their names 
and in how they work. The prompts that prompt the models to reason are being improved with the outcome of the reasoning process, and in  Buffer-of-thoughts  thought-templates are used that are derived from other reasoning processes.
\citet{wang2023metacognitive} study Metacognitive-prompting.  Inspired by Chain-of-thought and
Self-consistency, they create manually designed prompts to
increase the understanding of language models.
%
%
%
%
Another work, again inspired by Chain-of-thought and Self-consistency, connects psychology and LLMs. \citet{didolkar2024metacognitive} study metacognitive 
capabilities of LLMs in mathematical problem solving, both on GSM8K
and on the harder MATH problems \citep{hendrycks2021measuring}. 
First, the model is prompted to find a skill name for each problem instance
in the dataset. For 7000 instances of GSM8K, 500 skill names were found by the model. Next, these 500 names are clustered down to 22 skills. They find that by using the names of these 22 skills in Chain-of-thought-like 
prompts, more problems are solved than with standard
Chain-of-Thought/Self-consistency/Program-aided-language prompts. Examples of the 22 skill names are {\em multiplication-and-addition}, {\em basic-arithmetic}, {\em subtraction}, and {\em algebra}.
%
%
Interestingly, the authors find that the skill exemplar repository
that is trained on a strong model (GPT-4), also down-translates to a weak
model (GPT-3). The performance of the weaker model   benefits from
the skill-name-enhanced prompts.
%
%
\citet{begus2025large} propose a program for the study of metalinguistic abilities of LLMs. LLMs  struggle with analogies \citep{lewis2024evaluating,mitchell1990emergence} and analogical reasoning. Metacognitive reasoning with LLMs is still in its early
stages.

\subsection{LLMs  that Generate Computer code}\label{sec:proglang}
A persistent thread in the results in this survey is that LLMs are good at generating formal languages, such as PDDL, logic, math equations, and Python. The agentic approach in which LLMs are combined with external tools such as interpreters or debuggers often yields good performance \citep{plaat2025agentic}. LLMs can be used in different modes, which we will now discuss.

When LLMs are used in {\em direct} mode, a prompt is provided with an instruction, and an answer. This is the mode in which most casual users  use an LLM.

LLMs can also be used in {\em algorithmic} mode, or Chain-of-thought mode, where an extra element is added to the prompt that gives an example of which steps to take to arrive at the answer \citep{sel2023algorithm}. \citet{muennighoff2025s1} show how such test-time scaling can trade-off model size and training time.  \citet{lee2023supervised,wang2024transformers,schultz2024mastering} show that the transformer architecture can be trained such that search and reinforcement learning algorithms can be executed in-context. 

LLMs do not preserve state between calls, and in order to allow for a natural dialogue to occur, chatbots such as ChatGPT  insert the most recent history of preceding calls into each prompt before the model is called  \citep{brown2020language}.  For harder  strategic reasoning tasks 
more advanced state management solutions are necessary, that call the LLM in an explicit {\em external} mode \citep{yao2024tree,yao2022react}. Approaches such as Tree-of-thoughts use the LLM  in a stateless functional way, with the prompt containing the instructions, not unlike a natural language variant of a functional programming language such as LISP or Haskell.  

Finally, an LLM can be used in {\em code generation} mode. The prompt specifies a problem, and the LLM is asked to generate computer code (such as Python or PDDL), to circumvent the knowing-doing gap \citep{paglieri2024balrog}. The code may contain variables that manage the state correctly. The code can be executed directly, or first be checked and improved by a programmer. 
We expect that the use LLMs as code generators, and prompt-engineering as natural-language-programming, will continue to grow.

\subsection{Research Agenda}
At the end of this discussion, we list promising topics for future work.
Reasoning with LLMs is an active field of
research. It brings together elements of symbolic reasoning,
connectionism, natural language, autonomous agents,  affective reasoning \citep{broekens2023fine} and metacognition. 
%
First we discuss current topics for the field of LLM-reasoning itself,
then we discuss more general machine learning topics that are
important for progress in LLM-reasoning, and finally we discuss more longer term, fundamental topics.

Specific research topics for reasoning with LLMs are: 
\begin{itemize}
\item {\em In Context Reinforcement Learning}---
  Search control beyond greedy search
  is often implemented as an external reinforcement learning 
  algorithm. Is it possible to incorporate the control stage of the
  reasoning pipeline into one static prompt, for implicit reasoning or in-context reinforcement learning (ICRL)? Studies indicate that models can be trained for decision making \citep{lee2023supervised,demircan2024sparse}, but general LLMs struggle to perform ICRL well \citep{schultz2024mastering,paglieri2024balrog,ruoss2024lmact,van2025baba}.
\item {\em Code}---Progress in reasoning using formal languages and computer code has been
  quite promising. GitHub Copilot is a success. Further integration of
  LLM-reasoning with software engineering tools is a promising area of research that can have a large practical impact on how software is written.
\item {\em Grounding}---Reasoning in LLMs has been successfully applied in  
  autonomous agents, robotics, and games. A challenge is the grounding of the reasoning process in the
  environment. Retrieval augmented generative methods (RAG) can  help LLMs to actively find new information when
  the reasoning outcome is uncertain. RAG, search, and agentic LLMs, are also active areas of research \citep{dai2022promptagator,verberne2024zoekmachine,plaat2025agentic}. 

\end{itemize}
Generic topics in machine learning that also influence prompt-based reasoning
research are:
\begin{itemize}
\item {\em Benchmarks}---Progress in LLMs depends on the
  availability of the right
  benchmarks.  As
  the field has progressed beyond math word problems, other benchmarks  become prevalent, with more difficult and diverse tasks.  
\item {\em Faithfulness}---Our theoretical understanding of prompt-based
  reasoning with LLMs is incomplete. The research on faithfulness highlights one
  example of our lack of understanding. In general, more insight into implicit reasoning \citep{li2025implicit} and
  the working of multi-step in-context learning in LLMs is dearly needed.
\item {\em Smaller language models}---Efficiency is an important element for wide adoption of language models. Smaller models have many advantages over larger models: a smaller environmental footprint, more accessible to people with fewer computational resources, and the possibility to finetune them. Models may have fewer parameters, or the parameters may be quantized with fewer bits. Unfortunately, with a smaller number of parameters, the models also have less reasoning power. It is important for future work to investigate this trade-off. 
\end{itemize}
%
For longer term future work, the following more fundamental questions are important:
\begin{itemize}
\item {\em Symbolic and Connectionist Computation}---How can we further improve LLM-reasoning: how can LLMs benefit from  reasoning prompts based on the symbolic AI literature, and how can LLMs help ground symbolic reasoning in language? 

\item {\em Multimodal World-Models with Embodied Grounding}---How can an LLM maintain a unified, continually updated representation that integrates text, vision, audio, and sensorimotor feedback, enabling it to reason over events across modalities, and refine its world-model through closed perception–action loops in real-world environments?

\item {\em Norm-Sensitive and Value-Aligned Reasoning}---How do we represent, adapt, and audit diverse cultural norms and ethical values within the reasoning process, ensuring that each step of an LLM’s Chain-of-thought respects context-dependent moral constraints while remaining transparent and verifiable?

\item {\em Reasoning in other languages than English}---The majority of the research into LLM reasoning is for English. Although there are efforts in creating benchmarks to test LLM capabilities in other languages\footnote{\url{https://github.com/NaiveNeuron/awesome-multilingual-llm-benchmarks}}, these contain mainly NLP tasks such as question answering (QA) and natural language inference (NLI). In addition, challenges related to low-resource languages (languages for which very limited training data is available) have not been addressed yet. 

\item {\em Reasoning in specialized domains}---In an effort to guide development and evaluation of new methods, research in AI has a strong focus on benchmarking: standardized datasets with a limited set of problems that are realistic to evaluate. In the real world, however, reasoning problems also occur in more challenging contexts. A few examples are: legal reasoning for the interpretation of case law or writing contracts; scientific reasoning for advancing scientific fields; and medical reasoning for AI-assisted diagnostics. 
  
\end{itemize}

\section{Conclusion}\label{sec:conclusion}


When large language models of sufficient size are prompted with  examples, they can perform few-shot learning in-context, providing an answer at inference time, without retraining model parameters. 
Although they achieve good performance on language tasks, simple prompting methods do not perform well on reasoning tasks---tasks that humans typically solve in a step by step fashion. 
Chain-of-thought is an in-context prompting method to 
guide an LLM to ``think step by step.'' Chain-of-thought was originally developed for solving grade school math word problems. 
%
GSM8K, the most popular reasoning benchmarks in this survey, contains 8500 grade school
math word problems. With older LLMs such as GPT-3, reasoning approaches show an improvement of
20-50\% points over standard prompting methods. This success has spawned many new reasoning approaches. 
A potential problem in the Chain-of-thought method is that errors may accumulate over multiple reasoning steps. Prompting the LLM to reformulate problems in Python code (or another formal language) can  successfully  reduce the error rate. 
The success of reasoning methods has attracted more applications, and with them, benchmarks are diverging.
In the field of autonomous agents and robotic action,  good
performance has been achieved by grounding  reasoning answers  in the environment and the
physical constraints of robotic movement. 
%
%

Many current LLMs have  adopted Chain-of-thought approaches. In this survey we categorize the approaches on how they  generate, evaluate, and control the reasoning steps. Inference-time reasoning methods are also used for LLM finetuning.
Reinforcement learning with verifiable rewards (RLVR) and group relative policy optimization (GRPO) use inference-time reasoning results to augment finetuning for reasoning tasks.

For complex  tasks the number of reasoning steps that is  generated may be large. The size of the multi-step reasoning space can be controlled dynamically by external search or reinforcement learning algorithms, often in the form of a wrapper of Python code that generates LLM-prompts.
%
Many of the names of the approaches in this survey suggest a link to metacognition ({\em Reflexion, Self-refine, Self-improvement,  Inner-monologue})---the act of thinking about one's own thought process.
The first preliminary experiments of
language models that reason about their reasoning skills have appeared.

The field of reasoning with LLMs is quite new, and
theoretical understanding is lacking in 
important areas, such as faithful reasoning (models may sometimes find the right answer using an incorrect reasoning chain). 
LLMs hallucinate and suffer from bias, and their use poses  ethical and societal dangers. Self-verification methods have been developed to reduce error-accumulation, and retrieval augmentation methods (RAG) ground LLM output directly in sources such as Wikipedia.
However, ethical and societal dangers remain, especially since also reflection is not error-free. 
%
%
%
%

Although prompt-based learning allows few-shot learning at inference time, the computational needs of LLM pretraining and finetuning are still  high, hence the interest in small language models.  Reasoning skills that work in large models can often be distilled  to small models.


LLM-reasoning is an active field of research that shows great progress. 
Based on current limitations and open questions we provide a research agenda highlighting opportunities for further progress. Among the items mentioned are reinforcement learning for finetuning and self-reflection, LLM agents that generate code for external tools,  multimodal world models,  value-aligned reasoning, and  small language models.

\section*{Acknowledgements}
We thank the anonymous reviewers for their valuable   suggestions that have considerably improved the article. 
\bibliographystyle{plainnat}
\bibliography{resllm}

\end{document}